\documentclass{article}

\usepackage[preprint]{neurips_2026}

\usepackage[utf8]{inputenc} 
\usepackage[T1]{fontenc}    
\usepackage{url}            
\usepackage{nicefrac}       
\usepackage{microtype}      

\usepackage{times}
\usepackage{latexsym}
\usepackage{graphicx}
\usepackage{textcomp}
\usepackage[table]{xcolor}
\usepackage{physics}
\usepackage{mathdots}
\usepackage{algorithm}
\usepackage{algpseudocode}
\usepackage{subfigure}
\usepackage{pgfplots}
\usepackage{tikz}
\usetikzlibrary{patterns,arrows.meta,calc,positioning,fit,shapes.geometric}
\usepackage{multirow}
\usepackage{multicol}
\usepackage{array}
\pgfplotsset{compat=1.18}

\usepackage{sidecap}

\usepackage{booktabs}       
\usepackage{tcolorbox}
\tcbuselibrary{breakable,skins}  
\usepackage{newunicodechar}
\newunicodechar{─}{\rule[0.4ex]{0.9ex}{0.4pt}}
\newunicodechar{│}{\rule[-0.5ex]{0.4pt}{2ex}}
\newunicodechar{├}{\rule[-0.5ex]{0.4pt}{2ex}\hspace{-0.4pt}\rule[0.4ex]{0.7ex}{0.4pt}}
\newunicodechar{└}{\rule[0.4ex]{0.4pt}{1.1ex}\hspace{-0.4pt}\rule[0.4ex]{0.7ex}{0.4pt}}
\newunicodechar{→}{$\rightarrow$}
\usepackage{alltt}

\usepackage{amsmath,amssymb,amsfonts}
\usepackage{dsfont}

\usepackage{hyperref}
\hypersetup{
    colorlinks=true,
    linkcolor=blue,
    filecolor=magenta,
    urlcolor=cyan,
    pdftitle={SpecBlock: Block-Iterative Speculative Decoding with Dynamic Tree Drafting},
    pdfauthor={Weijie Shi, Qiang Xu, Fan Deng, Yaguang Wu, Jiarun Liu, Yehong Xu, Hao Chen, Jia Zhu, Jiajie Xu, Xiangjun Huang, Jian Yang, Xiaofang Zhou},
    pdfpagemode=FullScreen,
    }

\title{SpecBlock: Block-Iterative Speculative Decoding \\ with Dynamic Tree Drafting}

\author{%
  Weijie Shi$^{1}$ \quad Qiang Xu$^{2}$ \quad Fan Deng$^{2}$ \quad Yaguang Wu$^{2}$ \quad Jiarun Liu$^{2}$ \\
  Yehong Xu$^{1}$ \quad Hao Chen$^{1}$ \quad Jia Zhu$^{3}$ \quad Jiajie Xu$^{4}$ \quad Xiangjun Huang$^{2}$ \\
  Jian Yang$^{2}$ \quad Xiaofang Zhou$^{1}$ \\[0.6em]
  $^{1}$Hong Kong University of Science and Technology \quad $^{2}$MetaX \\
  $^{3}$Zhejiang Normal University \quad $^{4}$Soochow University \\
}

\begin{document}

\maketitle

\begin{abstract}
Speculative decoding accelerates LLM inference by drafting a tree of candidate continuations and verifying it in one target forward. Existing drafters fall into two camps with opposite weaknesses. Autoregressive drafters such as EAGLE-3 preserve dependence along each draft path but call the drafter once per tree depth, making drafting a non-trivial share of per-iteration latency. Parallel drafters cut drafter calls by predicting multiple future positions in one forward, but each position is predicted without seeing the others, producing paths the verifier rejects. In this paper, we propose SpecBlock, a block-iterative drafter that combines path dependence with cheap drafting. Each drafter forward produces $K$ dependent positions and we call this a block. The draft tree grows through repeated block expansions. Two mechanisms explicitly carry path dependence to keep later draft positions accurate. Within each block, a layer-wise shift carries the previous position's hidden state into every decoder layer. Across blocks, each new block can start from any position of the previous block, inheriting its hidden state to extend the path. To spend verifier budget where acceptance is likely, a co-trained rank head replaces the fixed top-$k$ tree by allocating per-position branching during drafting. To avoid training the drafter on prefixes it never produces at inference, a valid-prefix mask drops the loss at later positions once an earlier one is wrong. Beyond static drafting, a cost-aware bandit at deployment uses free verifier feedback to update the drafter selectively, only when the expected throughput gain exceeds the update cost. Experiments show that SpecBlock improves mean speedup by $8$--$13\%$ over EAGLE-3 at $44$--$52\%$ of its drafting cost, and cost-aware adaptation extends this lead to $11$--$19\%$.
\end{abstract}

\section{Introduction}
\label{sec:intro}

Since large language model (LLM) decoding is often limited by memory bandwidth, speculative decoding~\citep{leviathan2023fast,chen2023accelerating,miao2023specinfer,sun2023spectr,zhou2023distillspec} addresses this bottleneck by using a small draft model to predict multiple future tokens and letting the target model verify these candidates in parallel, allowing one target forward to accept multiple tokens and use the available compute capacity more fully.
Tree-based verification~\citep{miao2023specinfer,chen2024sequoia} further replaces a single drafted sequence with a draft tree, giving the verifier multiple alternatives at future positions and substantially increasing the acceptance length.
The gain from tree-based verification depends on a balance: the tree must cover continuations the target is likely to accept while keeping draft computation small enough that the saved target calls translate into net speedup.

\begin{figure}[t]
  \centering
  \includegraphics[width=\linewidth]{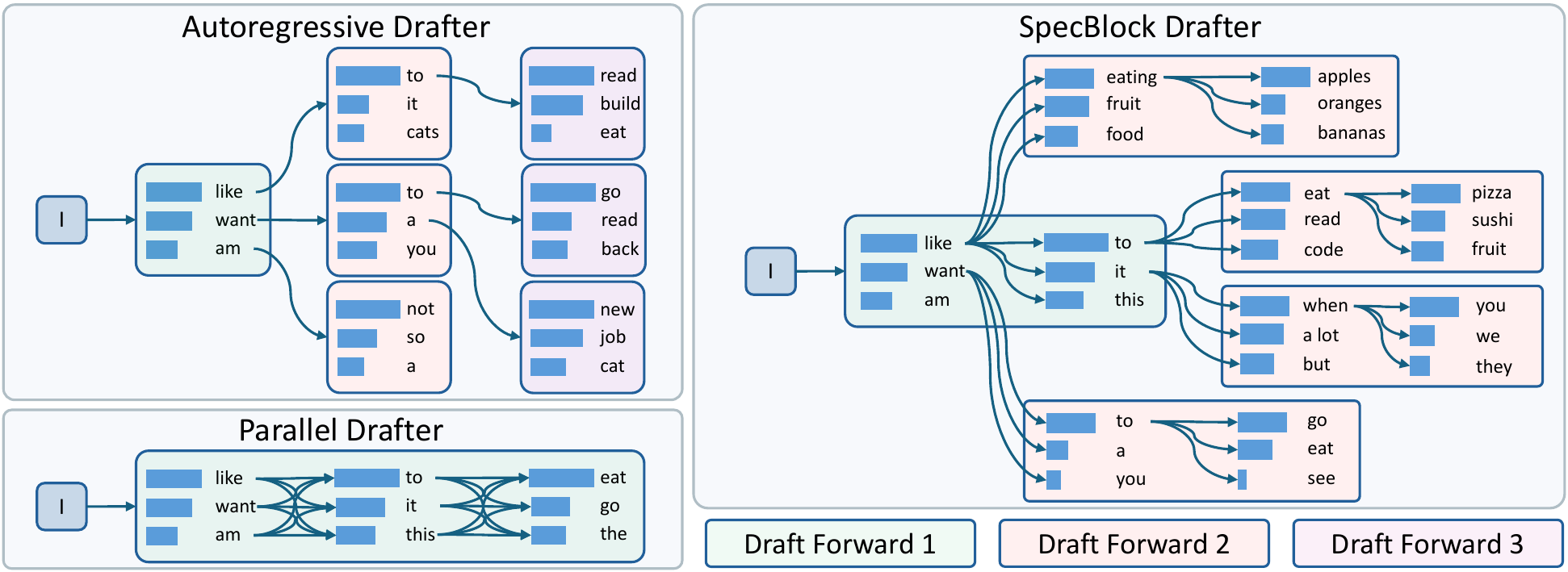}
  \caption{Three drafting paradigms. Autoregressive drafters (\textbf{left}) add one depth per drafter call. Parallel drafters (\textbf{bottom-left}) predict all depths independently in one call. SpecBlock (\textbf{right}) produces $K$ dependent positions per call and batches blocks from multiple starting positions into each subsequent call, growing the tree iteratively.}
  \label{fig:paradigms}
\end{figure}

Autoregressive drafters such as EAGLE-3~\citep{li2024eagle,li2024eagle2,li2025eagle3} grow the draft tree depth by depth. This preserves dependence along each draft path and can reach average acceptance lengths near 6, but each added tree depth still costs one more sequential drafter round. Although the drafter is small, each round is itself memory-bound, so the serial calls accumulate bursts of weight loading and consume close to $30\%$ of per-iteration latency on 8B-level target. Parallel drafters~\citep{cai2024medusa} reduce this overhead by proposing several future positions in one call, shrinking drafting to roughly $7\%$. However, once alternatives from different depths are combined into a draft tree, they form a large combinatorial space in which many paths are not coherent continuations, and the verifier wastes budget on them. This calls for a balance between the two camps, a drafter that both makes few drafter calls and preserves path coherence along each draft path, as illustrated in Figure~\ref{fig:paradigms}.

To realize this balance, we propose \textbf{SpecBlock}, a block-iterative drafter that treats each draft forward as producing a multi-token block and grows the draft tree through repeated block expansions. A generated tree node can serve as the starting point of a subsequent block, so one batched draft forward can extend multiple branches in parallel rather than expanding the tree depth by depth. Two mechanisms keep later positions accurate inside this construction by explicitly carrying dependence. Within each block, a layer-wise shift carries the previous position's hidden state into every decoder layer. Across blocks, each new block can continue from any position of the previous block, conditioned on that position's hidden state.

\textbf{Rank-guided tree construction.} Different draft positions deserve different amounts of branching, because the target token may sit at the top of the draft distribution at one position and far down at another. A co-trained rank head reads each position's hidden state and predicts how high the target token ranks in that position's draft distribution, expressed as a coarse bucket. This bucket sets the number of sibling alternatives at the position and decides whether the position starts a later block, so the tree is shaped on the fly during drafting rather than pruned afterwards.

\textbf{Valid-prefix curriculum learning.} An autoregressive drafter teacher-forces each step from a fresh ground-truth prefix, so every step's loss is supervised under the correct context. SpecBlock cannot do this, because its $K$ predictions are produced jointly in one forward and later positions read the actual earlier predictions instead of the ground-truth prefix. If an earlier prediction is wrong, the verifier rejects the entire path. Supervising later positions on the ground-truth target then only spends capacity on tokens the drafter will never commit. The valid-prefix mask therefore drops the loss at any later position once an earlier one on the same path is wrong.

\textbf{Cost-aware serving-time adaptation.} A small drafter trained offline cannot fit every domain, and acceptance length drops when the serving prompt distribution moves away from the training mix. The verifier already produces a free adaptation signal at every query, namely the target distribution at each rejected position, which is computed during verification at no extra cost. A cost-aware bandit reads this signal and decides whether to skip the update, update only the output heads, or update the full drafter, taking a non-skip action only when the expected throughput gain exceeds the update cost.

Experiments show that SpecBlock improves mean speedup by $8$--$13\%$ over EAGLE-3 at $44$--$52\%$ of its drafting cost. Cost-aware serving-time adaptation widens this advantage to $11$--$19\%$ on benchmarks with sufficient streaming queries.

\section{Related Work}
\label{sec:related}

\paragraph{Autoregressive drafters.}
Standard speculative decoding~\citep{leviathan2023fast,chen2023accelerating} establishes a lossless draft-then-verify framework, where a small drafter proposes a chain of future tokens that the target verifies in parallel. SpecInfer~\citep{miao2023specinfer} generalizes the chain into a token tree so one verification accepts the longest matching path among multiple candidates, lifting accepted length when the drafter is uncertain. The EAGLE family~\citep{li2024eagle,li2024eagle2,li2025eagle3} then trades drafter capacity for fidelity by autoregressing in the target model's feature space and growing dynamic trees from drafter confidence, with HASS~\citep{zhang2025hass} further aligning training and inference by simulating the drafter's own multi-step rollout. Other variants cut drafter overhead through distillation~\citep{zhou2023distillspec}, by reusing the target's own shallow layers~\citep{zhang2023draft,liu2024kangaroo}, or by retrieving cached continuations from a datastore~\citep{he2023rest,fu2024break}. All of these methods gain acceptance by propagating dependence one depth at a time, so each added depth still costs another drafter step.

\paragraph{Parallel and blockwise drafters.}
Parallel and blockwise drafters take the opposite trade-off and cut the drafter to one forward by predicting several future positions at once. Each position is predicted independently of the others, so the draft tree fails to capture the dependence between adjacent tokens, and its paths diverge from the target's continuation after the first few depths. Draft-head methods~\citep{stern2018blockwise,cai2024medusa} attach independent heads at fixed offsets. Mask-based drafters predict each future offset from a learnable mask token, with BiTA~\citep{lin2024bita}, ParallelSpec~\citep{xiao2024parallelspec}, and PARD~\citep{an2025pard} differing in how the masks are integrated, and DART~\citep{liang2026dart} layering a diffusion-style masked-prediction objective on top. Hydra~\citep{ankner2024hydra} re-introduces dependence by chaining the heads sequentially, each conditioning on the candidate continuation produced by earlier heads. Blockwise and semi-autoregressive variants instead enlarge the draft unit, through layer cascades~\citep{fasteagle2025}, semi-autoregressive block drafting~\citep{hu2024falcon,liu2024pearl}, or recurrent and block-mask architectures~\citep{kim2024bpd,gat2025sbd,cheng2024redrafter}, raising the average tokens per draft call. Falcon is the closest of these and also drafts semi-autoregressive blocks, but carries within-block dependence through stacked LSTM layers and relaxed-causal-mask attention that lets all positions inside a block see one another, and verifies a hand-crafted static decoding tree. SpecBlock instead enforces strict left-to-right within-block dependence through a per-layer hidden-state shift, and shapes the verifier tree dynamically through a co-trained rank head.

\paragraph{Tree construction.}
On top of an autoregressive drafter, the draft tree is shaped externally from drafter signals to decide which nodes the target verifies. Sequoia~\citep{chen2024sequoia} solves an offline DP over tree size and depth, while C2T~\citep{huo2025c2t}, OPT-Tree~\citep{wang2025opttree}, DySpec~\citep{xiong2025dyspec}, and TALON~\citep{liu2026talon} adapt the tree from drafter probability, confidence, or budget signals. SpecBlock instead integrates tree construction into the drafter through a rank head that sets per-position branching and which positions start later blocks.

\paragraph{Serving-time adaptation.}
Speculative drafters are kept small to make drafting cheap, which also makes them sensitive to serving-time distribution shifts that lower acceptance length. One line leaves drafter weights frozen and adapts only the speculation hyperparameters such as proposal length and tree size, either via a bandit over candidate configurations~\citep{hou2025banditspec} or via a learned threshold on per-position acceptance probability~\citep{huang2024specdec}. Another updates the drafter through verifier-feedback distillation~\citep{liu2024online}, with extensions reformulating the drafter as a self-speculative head trained with a KL-to-RL schedule on accepted-position rewards~\citep{qian2026dvi}, scheduling updates over time~\citep{park2026tide}, or integrating training tighter into the serving stack~\citep{wang2026aurora}, but each commits the trainable parameter subset ahead of time. Beyond when to update, SpecBlock makes which subset of drafter parameters to refresh a per-query decision, exposing a heads-versus-full-drafter action split where the rank head and \texttt{lm\_head} form a self-contained output-side pathway that can absorb output-level mistakes without touching the decoder.

\section{SpecBlock}
\label{sec:method}

Let $\mathcal{M}$ be the target model and $\mathcal{D}_\theta$ a small drafter that proposes a tree of candidate continuations for $\mathcal{M}$ to verify in one parallel forward. Speculative decoding throughput is the ratio $\Phi = \tau\,/\,(T_{\mathcal{M}} + T_{\mathcal{D}})$, where $\tau$ is the average length accepted per verifier call, $T_{\mathcal{M}}$ is the time of one target forward, and $T_{\mathcal{D}}$ is the cost of all drafter calls used to assemble that tree. Autoregressive drafters keep $\tau$ high but invoke the drafter once per tree depth, paying $T_{\mathcal{D}}$ proportional to depth. Parallel drafters collapse $T_{\mathcal{D}}$ to a single forward but lose $\tau$ because each future position is predicted without seeing the others.

SpecBlock improves $\Phi$ by sitting between these extremes: each drafter forward produces $K$ dependent positions, and the tree past depth $K$ is grown by re-using the drafter on a batch of starting points selected from earlier blocks. We further shape the tree's branching during drafting through a co-trained rank head, train the drafter under the prefix distribution it actually faces at inference, and refresh it selectively at serving time with a verifier-derived update signal.

\subsection{The drafter block}
\label{sec:method:arch}

Like prior speculative sampling methods, SpecBlock alternates between drafting and verification. The difference from EAGLE-3~\citep{li2025eagle3} lies in the drafting stage, where each drafter forward predicts $K$ consecutive positions in parallel as one block.


\begin{figure}[t]
  \centering
  \includegraphics[width=\linewidth]{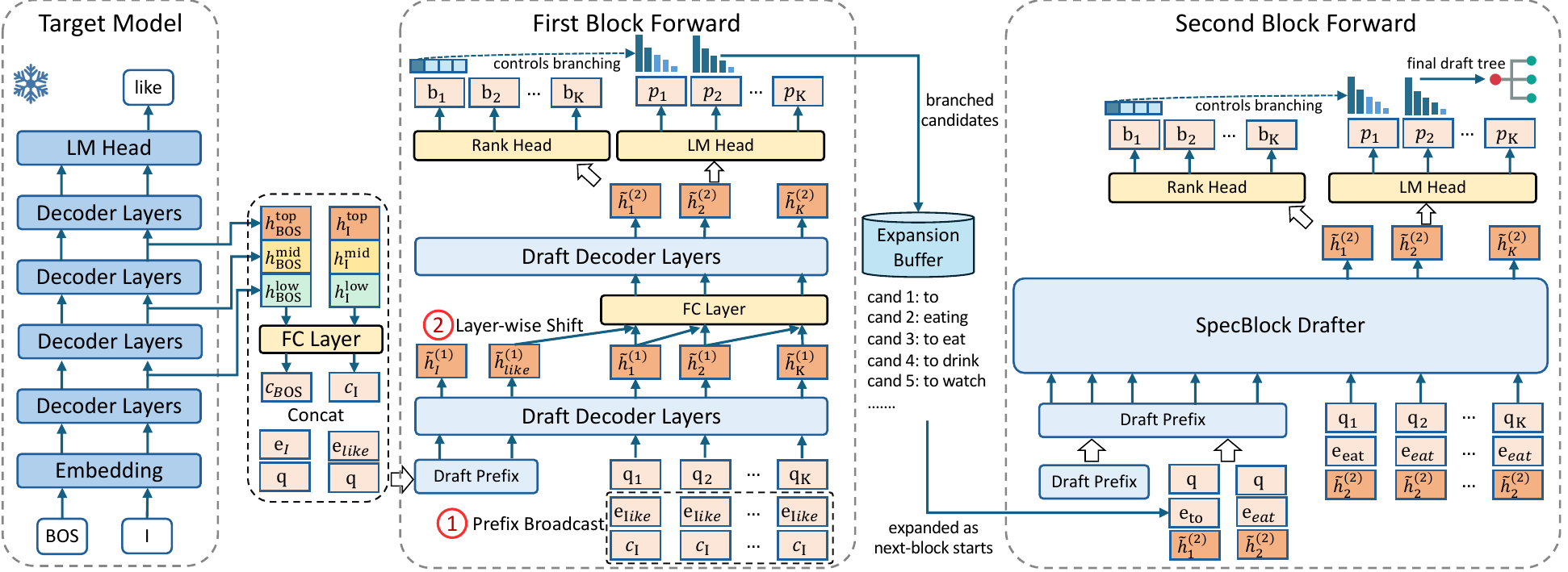}
  \caption{SpecBlock drafter architecture and block-iterative drafting.
    The first block (middle) fuses the target's multi-layer hidden state,
    the verified token, and $K$ position queries; prefix broadcast
    (\textcircled{\scriptsize 1}) shares the target features across the $K$
    positions, and a layer-wise shift (\textcircled{\scriptsize 2})
    propagates position $k{-}1$'s state into position $k$ between
    consecutive decoder layers. The lm head outputs draft distributions
    $p_k$ and the rank head outputs bucket labels $b_k$ that control
    per-position branching width. Branched candidates fill the expansion
    buffer and feed block 2 (right) as next-block starts.}
  \label{fig:arch}
\end{figure}

\paragraph{Block forward.}
Consider a draft block at the verified prefix's last position $t$, illustrated in Figure~\ref{fig:arch}. Following EAGLE-3, we build a context feature $c_t$ by concatenating the target model $\mathcal{M}$'s low-, mid-, and top-layer hidden states at position $t$ and projecting them to the drafter's dimension $d$ via a learned linear projection $W_{\text{cond}}$:
\begin{equation}
c_t \;=\; W_{\text{cond}}\,[\,h^{\text{low}}_t \,,\, h^{\text{mid}}_t \,,\, h^{\text{top}}_t\,].
\end{equation}
The other two inputs to the drafter are the embedding of the last committed token $x_t$ and $K$ learnable position queries $\mathbf{q}_1,\dots,\mathbf{q}_K$, one per draft depth. The three signals are normalized and fused into a per-position input via a learned linear projection $W_{\text{fuse}}$:
\begin{equation}
h^{(0)}_{t,k} \;=\; W_{\text{fuse}}\bigl[\,\mathrm{norm}(c_t) \,,\, \mathrm{norm}(\mathrm{embed}(x_t)) \,,\, \mathrm{norm}(\mathbf{q}_k)\,\bigr].
\end{equation}
A prefix broadcast ties the same $c_t$ and $\text{embed}(x_t)$ across positions so each one receives the prefix context directly. Only $\mathbf{q}_k$ varies across positions. This per-position input then passes jointly through the drafter's $L$ Transformer decoder layers, which match $\mathcal{M}$'s per-layer architecture, giving the last-layer state $h^{(L)}_{t,k}$ at each position. The lm\_head reads $h^{(L)}_{t,k}$ to produce the draft distribution $p_{t,k}$, and we cache $h^{(L)}_{t,k}$ for downstream blocks.

\paragraph{Within-block dependence.}
The $K$ positions are produced jointly. Any coherence along a draft path must therefore come from interactions inside the $L$ decoder layers. Cross-position causal attention restricts position $k$ to attend only to positions $\le k$ within the block, plus preceding blocks via cached key-value pairs, reproducing left-to-right dependence at the attention level. However, each attended position contributes only one weight in the softmax mixture, and that weight is diluted as the prefix grows, collapsing acceptance at deeper positions of a block. We therefore add a layer-wise shift between consecutive decoder layers that explicitly carries position $k{-}1$'s state into position $k$. This approximates in one forward the state propagation that EAGLE-3 obtains by running a separate drafter forward per position. Before entering layer $\ell{+}1$, position $k$'s state is concatenated with position $k{-}1$'s state from the same layer and projected back to $\mathbb{R}^d$ via a per-layer learned linear projection $W^{(\ell)}_{\text{shift}}$,
\begin{equation}
\tilde h^{(\ell)}_{t,k} \;=\; W^{(\ell)}_{\text{shift}}\bigl[\,h^{(\ell)}_{t,k} \,,\, h^{(\ell)}_{t,k-1}\,\bigr],
\end{equation}
with the convention $h^{(\ell)}_{t,0} = h^{(\ell)}_{t,1}$ at $k{=}1$. This recovers the dependence lost to attention dilution while staying within a single drafter forward.

\subsection{Rank-guided tree expansion}
\label{sec:method:rank}

A draft tree grows along two axes: depth, by chaining additional drafter forwards past the first block, and width, by attaching sibling alternatives at each position. The verifier budget along both axes should track the drafter's uncertainty. At an easy position one child suffices, while at a harder position the target sits several ranks deeper and the path is recovered only if at least one of several alternatives matches. A fixed branching factor either over-spends on easy positions or under-explores hard ones. A co-trained rank head coordinates both axes. Its bucket prediction at each position determines both the per-position branching width and whether the position starts a later block.

\paragraph{Rank head.}
The rank prediction needs features that reflect both the drafter's internal confidence and the shape of its output distribution. The rank head $g_\phi$ reads two such features at each position: the last-layer hidden state $h^{(L)}_{t,k} \in \mathbb{R}^d$, which carries the drafter's contextual representation, and a fixed 15-dimensional summary $\psi(p_{t,k})$ of the draft distribution $p_{t,k}$, detailed in Appendix~\ref{app:rank_head}. Both inputs are detached from the drafter's gradient via the stop-gradient operator $\mathrm{sg}(\cdot)$,
\begin{equation}
g_\phi\bigl(\bigl[\,\mathrm{sg}(h^{(L)}_{t,k}) \,,\, \mathrm{sg}(\psi(p_{t,k}))\,\bigr]\bigr) \;\in\; \{b_0, b_1, b_2, b_3\},
\end{equation}
so that the rank objective shapes the head's parameters but not the drafter trunk, leaving token prediction unaffected.

\paragraph{Bucket-driven branching.}
The optimal branching factor changes sharply with rank, not smoothly. Per-rank training samples are also highly imbalanced, with rank-1 dominating and distant ranks rare. We therefore collapse the rank prediction into four coarse buckets and assign each bucket a branching factor $b$, so position $k$ attaches the top-$b$ tokens of $p_{t,k}$ as siblings within its block. Confident positions attach few siblings since the target is already near the top of the drafter's distribution, while uncertain positions attach more siblings to widen the recovery window.

\paragraph{Cross-block iteration.}
Cross-block iteration re-invokes the drafter from positions whose rank-head bucket schedules them as next-block starts. These positions are batched into one drafter forward to produce $K$ further positions from each. The condition $c_t$ at each such point is no longer the target model's hidden state but the drafter's own cached $h^{(L)}_{t,k}$, which is already in $\mathbb{R}^d$ and bypasses $W_{\text{cond}}$. We use the drafter's self-produced features here because the target has not yet verified the position, so no target hidden state is available. We bound the chain at $M$ blocks, so the longest path in the tree reaches depth $M{\cdot}K$ at the cost of $M$ drafter forwards.

\subsection{Valid-prefix curriculum learning}
\label{sec:method:training}

The drafter and rank head should be trained under conditions consistent with inference. An autoregressive drafter teacher-forces each step on the ground-truth prefix, so every supervision signal sees a right-prefix context. SpecBlock cannot do the same. All $K$ positions of a block are produced jointly in one forward, with position $k$'s hidden state built from the drafter's own representations at earlier positions, so ground-truth tokens cannot be spliced in mid-forward. When an earlier prediction is wrong, later positions are supervised under a wrong-prefix context, which both interferes with right-prefix supervision and is wasted because the verifier truncates the path at the first deviation. We therefore mask both the draft loss and the rank-head loss on any path within the block that has deviated.

\paragraph{Valid-prefix mask.}
We define a binary mask $m_{t,k} \in \{0,1\}$ along each path of a block. The mask is initialized to $m_{t,1} = 1$ at the first position of every path. After each draft position, the mask updates by
\begin{equation}
m_{t,k+1} \;=\; m_{t,k} \,\cdot\, \mathds{1}\!\left[\arg\max p_{t,k} \,=\, y^\star_{t,k}\right],
\end{equation}
where $y^\star_{t,k}$ is the target token at the offset that draft position $k$ predicts. We compute position $k$'s draft loss only on the paths the mask still admits,
\begin{equation}
\mathcal{L}^{\text{draft}}_k \;=\; -\frac{1}{N_k}\sum_{t} m_{t,k} \sum_{v} p^\star_{t,k}(v)\,\log p_{t,k}(v),\qquad N_k \,=\, \sum_{t} m_{t,k},
\end{equation}
where $p^\star_{t,k}$ is the target's next-token distribution at the matching offset.

\paragraph{Rank-head supervision.}
For each training position we compute the target token's rank $r$ within $p_{t,k}$ and assign the bucket label by the rule $r{=}1 \!\mapsto\! b_0$, $r \in [2,4] \!\mapsto\! b_1$, $r \in [5,10] \!\mapsto\! b_2$, $r > 10 \!\mapsto\! b_3$. The rank head is supervised with cross-entropy against this label, masked by the same valid-prefix mask $m_{t,k}$.

\paragraph{Cross-block training.}
Inference past the first block conditions on the drafter's own cached hidden state rather than on the target's multi-layer feature. To expose the drafter to this shift during training, at each block boundary we sample a cut position $s$ uniformly from $\{1, \dots, K\}$, take the current block's last-layer hidden state $h^{(L)}_{t,s}$ as the next block's condition, and shift the ground-truth token sequence by $s$ positions as the next block's input. Uniform sampling of $s$ covers the full range of cross-block splits the rank head can produce at inference.

\paragraph{Total objective.}
The drafter is trained end-to-end with the sum of the $K$ per-position draft losses and the rank-head cross-entropy $\mathcal{L}^{\text{rank}}$, both masked by $m_{t,k}$,
\begin{equation}
\mathcal{L} \;=\; \sum_{k=1}^K \mathcal{L}^{\text{draft}}_k \;+\; \mathcal{L}^{\text{rank}}.
\end{equation}

\subsection{Cost-aware serving-time adaptation}
\label{sec:method:adapt}

The training procedure above yields a fixed drafter, but accepted length degrades when the serving prompt distribution shifts. Refreshing the drafter at serving time can restore the lost accepted length, but each backward roughly costs as much as one target forward, so an indiscriminate schedule negates the throughput it tries to protect. We therefore answer two questions per query: whether to update, and which parameters to update. The verifier's output provides a free signal for the first, and the drafter's modular architecture provides the action structure for the second.

\paragraph{Verifier-derived update signal.}
The drafter's distribution and the target's chosen token are both available at every rejected position, so reading them requires no extra work. For each rejected position $k$ on a verified path, let $r_{t,k} \in [0,1]$ be the drafter's probability of the target's chosen token. We aggregate these into a query-level signal
\begin{equation}
s \;=\; \sum_{k \,\in\, \text{rejected}} (1 - r_{t,k}),
\end{equation}
which is large when the drafter is far from the target's choices at multiple rejected positions and small when the two are nearly aligned.

\paragraph{Action set and per-query selection.}
A bandit selects per query among three actions, each addressing a different drafter-error mode: skip when the drafter is already well-calibrated, head-only when sound internal representations are mis-mapped by the lm\_head and rank head, and full update when the decoder trunk itself is mismatched. Let $s_{\text{trig}}$ denote the value of $s$ at the query that triggered an update, and let $v_{\text{head}}$ and $v_{\text{full}}$ each be an exponentially-weighted moving average (EWMA) of the realized reward $\Delta_{\text{tp}}^{\text{observed}}/s_{\text{trig}}$. Because throughput is measured over the interval that follows an update, $v_{\text{action}}$ already nets out the update's own cost. At query time we predict the net throughput gain of each non-skip action as
\begin{equation}
\widehat{\Delta}_{\text{tp}}(\text{action}) \;=\; s \cdot v_{\text{action}}.
\end{equation}
We pick the action with the largest predicted gain, and we skip if both predictions are non-positive. The measurement interval spans the $N$ queries following an update, and at its close the EWMA is revised as
\begin{equation}
v_{\text{action}} \;\leftarrow\; (1-\alpha)\,v_{\text{action}} \;+\; \alpha\,\frac{\Delta_{\text{tp}}^{\text{observed}}}{s_{\text{trig}}},
\end{equation}
with $\alpha = 0.10$. The revision is skipped when $s_{\text{trig}} < s_{\min}$, since dividing by a small $s_{\text{trig}}$ amplifies noise in the throughput estimate.

\paragraph{Asynchronous updates and drift control.}
The drafter is held in two copies at serving time, $\theta_{\text{inf}}$ and $\theta_{\text{train}}$. The bandit's chosen update is applied to $\theta_{\text{train}}$ on a separate stream, so drafting on $\theta_{\text{inf}}$ is not blocked. The trained copy is periodically copied back into $\theta_{\text{inf}}$. To prevent the drafter from drifting away from the pre-deployment distribution under repeated updates, we add a KL penalty $\lambda\,\mathrm{KL}(p_\theta \,\|\, p_{\theta_0})$ against the pre-deployment drafter $\theta_0$ to the per-update objective, with $\lambda = 0.01$. As a fail-safe we additionally monitor accepted length over a sliding window of three consecutive serving intervals, and if it decreases monotonically across all three we revert $\theta_{\text{inf}}$ to the most recent good checkpoint and reset the bandit's value estimates.

These four mechanisms together target the throughput formula $\Phi = \tau / (T_{\mathcal{M}} + T_{\mathcal{D}})$ from complementary angles. The block forward and the layer-wise shift cap $T_{\mathcal{D}}$ while preserving $\tau$ within a block. Rank-guided branching and valid-prefix training jointly lift $\tau$ at inference. Serving-time adaptation guards $\tau$ against deployment shift.

\section{Experiments}
\label{sec:experiments}

\subsection{Setup}

We evaluate SpecBlock on three target models, Llama-3.1-8B-Instruct~\citep{dubey2024llama}, Qwen3-8B, and Qwen3-32B~\citep{yang2025qwen3}. The drafter for each target is a stack of $L{=}2$ Transformer decoder layers matching the target's per-layer architecture, with per-block depth $K{=}4$ and $M{=}2$ blocks per inference iteration, materializing a verifier tree of up to $60$ nodes. We measure on a single NVIDIA A100-80GB GPU at batch size 1 with temperature 0 and 1.0, with cost-aware adaptation in its single-GPU variant unless noted. We report speedup (Spd) over vanilla autoregressive decoding, throughput $\Phi$ in tokens per second, accepted length $\tau$ averaged over verifier calls, and drafting cost $T_\mathcal{D}\%$ as the share of per-iteration latency spent on drafter forwards.
\paragraph{Baselines.}
We compare against representative drafters of several kinds, all reproduced under the same target model, attention backend, and tree-size budget. Vanilla decoding without speculation anchors the speedup. Standard speculative sampling (SpS)~\citep{leviathan2023fast,chen2023accelerating} uses a pretrained smaller model from the same family as the drafter, with no further training. We pair Llama-3.1-8B with Llama-3.2-1B, and the Qwen3 targets with Qwen3-0.6B. Among autoregressive drafters we use EAGLE-3~\citep{li2025eagle3}. Among parallel drafters we use Medusa~\citep{cai2024medusa} and ParallelSpec~\citep{xiao2024parallelspec}. Falcon~\citep{hu2024falcon} is the closest blockwise prior work. For online verifier-feedback adaptation, Online Speculative Decoding (OSD)~\citep{liu2024online} is instantiated on the SpecBlock drafter as SpecBlock+OSD.

\paragraph{Training.}
All trainable drafters are trained on prompts from UltraChat-200K\footnote{\url{https://huggingface.co/datasets/HuggingFaceH4/ultrachat_200k}}~\citep{ding2023enhancing} and ShareGPT,\footnote{\url{https://huggingface.co/datasets/Aeala/ShareGPT_Vicuna_unfiltered}} with answers regenerated by the target model so that the training distribution matches what the target actually emits at inference. Training runs for 20 epochs with AdamW (learning rate $5{\times}10^{-5}$, cosine schedule, gradient clip $0.5$). For SpecBlock, the valid-prefix mask and the cross-block training procedure of \S\ref{sec:method:training} are used throughout, and the rank head is enabled after the first $2{,}000$ update steps to let the drafter trunk reach a stable distribution before bucket supervision.

\paragraph{Evaluation tasks.}
We evaluate on six benchmarks spanning conversation, code, competition math, instruction following, question answering, and translation: MT-Bench~\citep{zheng2023judging}, HumanEval~\citep{chen2021evaluating}, MATH-500~\citep{hendrycks2021math}, Alpaca~\citep{alpaca}, Natural Questions (NQ)~\citep{kwiatkowski2019nq}, and WMT-23~\citep{kocmi2023wmt}.

\subsection{Main results}

Table~\ref{tab:main_results} reports per-benchmark speedup over vanilla decoding and accepted length $\tau$ on three target models at A100-80GB, batch size 1. Among static drafters, SpecBlock improves mean speedup over EAGLE-3 by $8$--$13\%$ across all six configurations. Cost-aware adapt further lifts speedup over the always-update SpecBlock+OSD by $2$--$4\%$ on the four benchmarks where the bandit engages. HumanEval at 164 prompts and MT-Bench at 80 prompts are too short for the acceptance gain to amortize the backward cost of adaptation, so we do not evaluate adapt on them. Target-verifier time is essentially constant across these methods at about $37$ ms per iteration on Llama-3.1-8B, so throughput reduces to how each drafter trades $\tau$ against drafter cost $T_\mathcal{D}$.

\begin{table}[t]
\centering
\caption{Speedup (Spd) over vanilla decoding and average accepted length $\tau$ per benchmark at A100-80GB, batch size 1, under HuggingFace Transformers. ``$T_\mathcal{D}\%$'' is the per-method drafting-cost share. Bold indicates the best speedup within each model group. SpecBlock+adapt is SpecBlock with cost-aware serving-time adaptation.}
\label{tab:main_results}
\setlength{\tabcolsep}{3pt}
\resizebox{\textwidth}{!}{%
\begin{tabular}{cl cccccccccccc cc c}
\toprule
& & \multicolumn{2}{c}{HumanEval} & \multicolumn{2}{c}{MATH-500} & \multicolumn{2}{c}{Alpaca} & \multicolumn{2}{c}{NQ} & \multicolumn{2}{c}{MT-Bench} & \multicolumn{2}{c}{WMT-23} & \multicolumn{2}{c}{Mean} & \\
\cmidrule(lr){3-4} \cmidrule(lr){5-6} \cmidrule(lr){7-8} \cmidrule(lr){9-10} \cmidrule(lr){11-12} \cmidrule(lr){13-14} \cmidrule(lr){15-16}
Model & Method & Spd & $\tau$ & Spd & $\tau$ & Spd & $\tau$ & Spd & $\tau$ & Spd & $\tau$ & Spd & $\tau$ & Spd & $\tau$ & $T_\mathcal{D}\%$ \\
\midrule
\multicolumn{17}{c}{\textit{Temperature = 0}} \\
\midrule
\multirow{8}{*}{Llama-3.1-8B}  & SpS          & $1.55\times$ & 2.91 & $1.17\times$ & 2.29 & $1.48\times$ & 2.70 & $1.06\times$ & 2.09 & $1.38\times$ & 2.45 & $0.99\times$ & 1.92 & $1.27\times$ & 2.39 & 38 \\
                         & Medusa       & $2.16\times$ & 2.70 & $1.53\times$ & 2.30 & $2.01\times$ & 2.44 & $1.44\times$ & 2.00 & $1.88\times$ & 2.59 & $1.34\times$ & 2.05 & $1.73\times$ & 2.35 &  6 \\
                         & ParallelSpec & $2.50\times$ & 3.16 & $1.89\times$ & 2.64 & $2.39\times$ & 3.28 & $1.78\times$ & 2.63 & $2.36\times$ & 3.21 & $1.66\times$ & 2.30 & $2.10\times$ & 2.87 &  8 \\
                         & Falcon       & $3.06\times$ & 4.69 & $2.30\times$ & 3.74 & $2.86\times$ & 4.23 & $2.12\times$ & 3.38 & $2.66\times$ & 4.31 & $1.90\times$ & 3.33 & $2.48\times$ & 3.95 & 26 \\
                         & EAGLE-3      & $3.59\times$ & 6.98 & $2.66\times$ & 5.93 & $3.35\times$ & 6.27 & $2.42\times$ & 5.39 & $\mathbf{3.22\times}$ & 6.16 & $2.28\times$ & 4.60 & $2.92\times$ & 5.89 & 31 \\
                         & \cellcolor{cyan!8}SpecBlock & \cellcolor{cyan!8}$\mathbf{3.92\times}$ & \cellcolor{cyan!8}5.16 & \cellcolor{cyan!8}$\mathbf{3.07\times}$ & \cellcolor{cyan!8}4.03 & \cellcolor{cyan!8}$\mathbf{3.40\times}$ & \cellcolor{cyan!8}4.66 & \cellcolor{cyan!8}$\mathbf{3.00\times}$ & \cellcolor{cyan!8}4.40 & \cellcolor{cyan!8}$3.10\times$ & \cellcolor{cyan!8}4.46 & \cellcolor{cyan!8}$\mathbf{2.79\times}$ & \cellcolor{cyan!8}3.75 & \cellcolor{cyan!8}$\mathbf{3.21\times}$ & \cellcolor{cyan!8}4.41 & \cellcolor{cyan!8}16 \\
\cmidrule(lr){2-17}
                         & SpecBlock+OSD & --- & --- & $3.10\times$ & 4.24 & $3.43\times$ & 4.74 & $3.25\times$ & 5.41 & --- & --- & $2.80\times$ & 3.85 & $3.14\times$ & 4.56 & 18 \\
                         & \cellcolor{cyan!8}SpecBlock+adapt & \cellcolor{cyan!8}--- & \cellcolor{cyan!8}--- & \cellcolor{cyan!8}$\mathbf{3.14\times}$ & \cellcolor{cyan!8}4.23 & \cellcolor{cyan!8}$\mathbf{3.47\times}$ & \cellcolor{cyan!8}4.72 & \cellcolor{cyan!8}$\mathbf{3.51\times}$ & \cellcolor{cyan!8}5.41 & \cellcolor{cyan!8}--- & \cellcolor{cyan!8}--- & \cellcolor{cyan!8}$\mathbf{2.81\times}$ & \cellcolor{cyan!8}3.81 & \cellcolor{cyan!8}$\mathbf{3.24\times}$ & \cellcolor{cyan!8}4.54 & \cellcolor{cyan!8}15 \\
\midrule
\multirow{4}{*}{Qwen3-8B}   & EAGLE-3      & $2.45\times$ & 4.53 & $1.94\times$ & 4.46 & $2.35\times$ & 4.29 & $\mathbf{2.50\times}$ & 4.20 & $\mathbf{2.40\times}$ & 4.23 & $1.81\times$ & 3.12 & $2.24\times$ & 4.14 & 29 \\
                         & \cellcolor{cyan!8}SpecBlock & \cellcolor{cyan!8}$\mathbf{2.50\times}$ & \cellcolor{cyan!8}3.59 & \cellcolor{cyan!8}$\mathbf{2.53\times}$ & \cellcolor{cyan!8}3.71 & \cellcolor{cyan!8}$\mathbf{2.58\times}$ & \cellcolor{cyan!8}3.73 & \cellcolor{cyan!8}$2.26\times$ & \cellcolor{cyan!8}3.21 & \cellcolor{cyan!8}$2.33\times$ & \cellcolor{cyan!8}3.32 & \cellcolor{cyan!8}$\mathbf{2.30\times}$ & \cellcolor{cyan!8}3.28 & \cellcolor{cyan!8}$\mathbf{2.42\times}$ & \cellcolor{cyan!8}3.47 & \cellcolor{cyan!8}14 \\
\cmidrule(lr){2-17}
                         & SpecBlock+OSD & --- & --- & $2.57\times$ & 3.92 & $2.62\times$ & 3.80 & $2.49\times$ & 3.75 & --- & --- & $2.32\times$ & 3.37 & $2.50\times$ & 3.71 & 19 \\
                         & \cellcolor{cyan!8}SpecBlock+adapt & \cellcolor{cyan!8}--- & \cellcolor{cyan!8}--- & \cellcolor{cyan!8}$\mathbf{2.61\times}$ & \cellcolor{cyan!8}3.90 & \cellcolor{cyan!8}$\mathbf{2.64\times}$ & \cellcolor{cyan!8}3.78 & \cellcolor{cyan!8}$\mathbf{2.69\times}$ & \cellcolor{cyan!8}3.75 & \cellcolor{cyan!8}--- & \cellcolor{cyan!8}--- & \cellcolor{cyan!8}$\mathbf{2.34\times}$ & \cellcolor{cyan!8}3.36 & \cellcolor{cyan!8}$\mathbf{2.56\times}$ & \cellcolor{cyan!8}3.70 & \cellcolor{cyan!8}17 \\
\midrule
\multirow{4}{*}{Qwen3-32B}  & EAGLE-3      & $2.54\times$ & 4.36 & $1.82\times$ & 4.27 & $2.28\times$ & 4.09 & $\mathbf{2.38\times}$ & 3.96 & $2.18\times$ & 4.09 & $1.71\times$ & 2.96 & $2.15\times$ & 3.96 & 24 \\
                         & \cellcolor{cyan!8}SpecBlock & \cellcolor{cyan!8}$\mathbf{2.61\times}$ & \cellcolor{cyan!8}3.47 & \cellcolor{cyan!8}$\mathbf{2.48\times}$ & \cellcolor{cyan!8}3.53 & \cellcolor{cyan!8}$\mathbf{2.48\times}$ & \cellcolor{cyan!8}3.54 & \cellcolor{cyan!8}$2.17\times$ & \cellcolor{cyan!8}3.07 & \cellcolor{cyan!8}$\mathbf{2.21\times}$ & \cellcolor{cyan!8}3.21 & \cellcolor{cyan!8}$\mathbf{2.20\times}$ & \cellcolor{cyan!8}3.15 & \cellcolor{cyan!8}$\mathbf{2.37\times}$ & \cellcolor{cyan!8}3.33 & \cellcolor{cyan!8}11 \\
\cmidrule(lr){2-17}
                         & SpecBlock+OSD & --- & --- & $2.51\times$ & 3.72 & $2.50\times$ & 3.62 & $2.36\times$ & 3.61 & --- & --- & $2.24\times$ & 3.24 & $2.40\times$ & 3.55 & 15 \\
                         & \cellcolor{cyan!8}SpecBlock+adapt & \cellcolor{cyan!8}--- & \cellcolor{cyan!8}--- & \cellcolor{cyan!8}$\mathbf{2.55\times}$ & \cellcolor{cyan!8}3.73 & \cellcolor{cyan!8}$\mathbf{2.51\times}$ & \cellcolor{cyan!8}3.61 & \cellcolor{cyan!8}$\mathbf{2.57\times}$ & \cellcolor{cyan!8}3.61 & \cellcolor{cyan!8}--- & \cellcolor{cyan!8}--- & \cellcolor{cyan!8}$\mathbf{2.24\times}$ & \cellcolor{cyan!8}3.22 & \cellcolor{cyan!8}$\mathbf{2.47\times}$ & \cellcolor{cyan!8}3.54 & \cellcolor{cyan!8}13 \\
\midrule
\multicolumn{17}{c}{\textit{Temperature = 1.0}} \\
\midrule
\multirow{8}{*}{Llama-3.1-8B}  & SpS          & $1.33\times$ & 2.32 & $0.58\times$ & 1.87 & $1.06\times$ & 2.33 & $0.70\times$ & 1.76 & $0.63\times$ & 2.01 & $0.82\times$ & 1.56 & $0.85\times$ & 1.97 & 40 \\
                         & Medusa       & $1.85\times$ & 2.20 & $0.83\times$ & 1.90 & $1.50\times$ & 2.07 & $1.01\times$ & 1.67 & $0.90\times$ & 2.13 & $1.10\times$ & 1.63 & $1.20\times$ & 1.93 &  7 \\
                         & ParallelSpec & $2.27\times$ & 2.65 & $0.97\times$ & 2.18 & $1.78\times$ & 2.85 & $1.20\times$ & 2.20 & $1.06\times$ & 2.67 & $1.33\times$ & 1.84 & $1.43\times$ & 2.40 &  9 \\
                         & Falcon       & $2.64\times$ & 3.86 & $1.15\times$ & 3.01 & $2.11\times$ & 3.64 & $1.44\times$ & 2.77 & $1.27\times$ & 3.55 & $1.51\times$ & 2.83 & $1.69\times$ & 3.28 & 27 \\
                         & EAGLE-3      & $2.98\times$ & 6.09 & $1.32\times$ & 4.31 & $2.36\times$ & 5.17 & $1.66\times$ & 4.27 & $1.48\times$ & 4.25 & $1.81\times$ & 3.94 & $1.94\times$ & 4.67 & 30 \\
                         & \cellcolor{cyan!8}SpecBlock & \cellcolor{cyan!8}$\mathbf{3.21\times}$ & \cellcolor{cyan!8}4.62 & \cellcolor{cyan!8}$\mathbf{1.74\times}$ & \cellcolor{cyan!8}3.24 & \cellcolor{cyan!8}$\mathbf{2.57\times}$ & \cellcolor{cyan!8}4.02 & \cellcolor{cyan!8}$\mathbf{1.92\times}$ & \cellcolor{cyan!8}3.60 & \cellcolor{cyan!8}$\mathbf{1.50\times}$ & \cellcolor{cyan!8}3.46 & \cellcolor{cyan!8}$\mathbf{2.31\times}$ & \cellcolor{cyan!8}3.41 & \cellcolor{cyan!8}$\mathbf{2.20\times}$ & \cellcolor{cyan!8}3.73 & \cellcolor{cyan!8}14 \\
\cmidrule(lr){2-17}
                         & SpecBlock+OSD & --- & --- & $1.77\times$ & 3.38 & $2.66\times$ & 4.25 & $2.17\times$ & 4.59 & --- & --- & $2.36\times$ & 3.59 & $2.24\times$ & 3.95 & 17 \\
                         & \cellcolor{cyan!8}SpecBlock+adapt & \cellcolor{cyan!8}--- & \cellcolor{cyan!8}--- & \cellcolor{cyan!8}$\mathbf{1.78\times}$ & \cellcolor{cyan!8}3.31 & \cellcolor{cyan!8}$\mathbf{2.72\times}$ & \cellcolor{cyan!8}4.22 & \cellcolor{cyan!8}$\mathbf{2.39\times}$ & \cellcolor{cyan!8}4.57 & \cellcolor{cyan!8}--- & \cellcolor{cyan!8}--- & \cellcolor{cyan!8}$\mathbf{2.36\times}$ & \cellcolor{cyan!8}3.57 & \cellcolor{cyan!8}$\mathbf{2.31\times}$ & \cellcolor{cyan!8}3.92 & \cellcolor{cyan!8}15 \\
\midrule
\multirow{4}{*}{Qwen3-8B}   & EAGLE-3      & $2.32\times$ & 4.34 & $2.18\times$ & 4.28 & $1.94\times$ & 4.14 & $\mathbf{1.94\times}$ & 4.09 & $1.93\times$ & 3.93 & $1.65\times$ & 3.10 & $1.99\times$ & 3.98 & 29 \\
                         & \cellcolor{cyan!8}SpecBlock & \cellcolor{cyan!8}$\mathbf{2.44\times}$ & \cellcolor{cyan!8}3.53 & \cellcolor{cyan!8}$\mathbf{2.60\times}$ & \cellcolor{cyan!8}3.63 & \cellcolor{cyan!8}$\mathbf{2.06\times}$ & \cellcolor{cyan!8}3.59 & \cellcolor{cyan!8}$1.81\times$ & \cellcolor{cyan!8}3.06 & \cellcolor{cyan!8}$\mathbf{2.06\times}$ & \cellcolor{cyan!8}3.19 & \cellcolor{cyan!8}$\mathbf{2.11\times}$ & \cellcolor{cyan!8}3.19 & \cellcolor{cyan!8}$\mathbf{2.18\times}$ & \cellcolor{cyan!8}3.37 & \cellcolor{cyan!8}14 \\
\cmidrule(lr){2-17}
                         & SpecBlock+OSD & --- & --- & $2.64\times$ & 3.83 & $2.14\times$ & 3.72 & $2.06\times$ & 3.55 & --- & --- & $2.16\times$ & 3.34 & $2.25\times$ & 3.61 & 19 \\
                         & \cellcolor{cyan!8}SpecBlock+adapt & \cellcolor{cyan!8}--- & \cellcolor{cyan!8}--- & \cellcolor{cyan!8}$\mathbf{2.67\times}$ & \cellcolor{cyan!8}3.79 & \cellcolor{cyan!8}$\mathbf{2.19\times}$ & \cellcolor{cyan!8}3.69 & \cellcolor{cyan!8}$\mathbf{2.27\times}$ & \cellcolor{cyan!8}3.55 & \cellcolor{cyan!8}--- & \cellcolor{cyan!8}--- & \cellcolor{cyan!8}$\mathbf{2.19\times}$ & \cellcolor{cyan!8}3.31 & \cellcolor{cyan!8}$\mathbf{2.33\times}$ & \cellcolor{cyan!8}3.59 & \cellcolor{cyan!8}18 \\
\midrule
\multirow{4}{*}{Qwen3-32B}  & EAGLE-3      & $2.17\times$ & 4.21 & $2.02\times$ & 4.06 & $1.84\times$ & 3.94 & $\mathbf{1.87\times}$ & 3.95 & $1.85\times$ & 3.82 & $1.55\times$ & 2.95 & $1.88\times$ & 3.82 & 25 \\
                         & \cellcolor{cyan!8}SpecBlock & \cellcolor{cyan!8}$\mathbf{2.26\times}$ & \cellcolor{cyan!8}3.43 & \cellcolor{cyan!8}$\mathbf{2.48\times}$ & \cellcolor{cyan!8}3.48 & \cellcolor{cyan!8}$\mathbf{1.94\times}$ & \cellcolor{cyan!8}3.42 & \cellcolor{cyan!8}$1.75\times$ & \cellcolor{cyan!8}2.92 & \cellcolor{cyan!8}$\mathbf{1.99\times}$ & \cellcolor{cyan!8}3.09 & \cellcolor{cyan!8}$\mathbf{2.07\times}$ & \cellcolor{cyan!8}3.10 & \cellcolor{cyan!8}$\mathbf{2.07\times}$ & \cellcolor{cyan!8}3.24 & \cellcolor{cyan!8}11 \\
\cmidrule(lr){2-17}
                         & SpecBlock+OSD & --- & --- & $2.51\times$ & 3.67 & $2.00\times$ & 3.55 & $1.99\times$ & 3.41 & --- & --- & $2.10\times$ & 3.23 & $2.15\times$ & 3.47 & 14 \\
                         & \cellcolor{cyan!8}SpecBlock+adapt & \cellcolor{cyan!8}--- & \cellcolor{cyan!8}--- & \cellcolor{cyan!8}$\mathbf{2.55\times}$ & \cellcolor{cyan!8}3.64 & \cellcolor{cyan!8}$\mathbf{2.06\times}$ & \cellcolor{cyan!8}3.53 & \cellcolor{cyan!8}$\mathbf{2.19\times}$ & \cellcolor{cyan!8}3.40 & \cellcolor{cyan!8}--- & \cellcolor{cyan!8}--- & \cellcolor{cyan!8}$\mathbf{2.14\times}$ & \cellcolor{cyan!8}3.22 & \cellcolor{cyan!8}$\mathbf{2.24\times}$ & \cellcolor{cyan!8}3.45 & \cellcolor{cyan!8}13 \\
\bottomrule
\end{tabular}%
}
\end{table}

Among static drafters, EAGLE-3 reaches the highest $\tau{=}5.89$ through seven sequential drafter calls but pays a $31\%$ drafter share, while parallel drafters such as Medusa and ParallelSpec cut this to $6$--$9\%$ at the cost of capping $\tau$ at around $2$ to $3$. SpecBlock condenses drafting into two block forwards, dropping drafter time from $17$\,ms to $7$\,ms while the layer-wise shift retains $\tau{=}4.41$ on Llama-3.1-8B, only $1.48$ tokens below EAGLE-3 despite the much shorter chain. Cost-aware adapt and SpecBlock+OSD reach essentially the same $\tau$, within $0.02$ tokens, but SpecBlock+adapt holds $T_\mathcal{D}\%$ at $13$--$18$ versus OSD's $14$--$19$: the bandit skips weak signals and routes most updates to the head-only action, whose backward over the lm\_head and rank head is more than an order of magnitude cheaper than a full-drafter backward. At $T{=}0$, single-domain benchmarks such as NQ and MATH-500 lift $\tau$ by $0.2$ to $1.0$ as streaming queries share a stable target, while mixed-instruction Alpaca lifts only $0.05$ to $0.07$ because gradients across instruction types partially cancel.

The trade-off strengthens on larger targets: drafter share falls from $31\%$ to $24\%$ for EAGLE-3 between Llama-3.1-8B and Qwen3-32B, and from $16\%$ to $11\%$ for SpecBlock, narrowing SpecBlock's relative drafting cost from $52\%$ to $46\%$ of EAGLE-3's. Sampling at $T{=}1.0$ reduces $\tau$ across all methods because rejection sampling against a high-entropy target accepts fewer draft tokens, but the relative ordering is unchanged.

\subsection{Ablations}

To evaluate each design component, Table~\ref{tab:ablations} removes one component per row on Llama-3.1-8B at $T{=}0$, with the base group on the six benchmarks and the adaptation group on the four where the bandit engages. Base rows ablate the prefix broadcast, the layer-wise shift, the valid-prefix curriculum, and the rank-guided branching, with the last replaced by a uniform fixed-$k$ tree at the same node budget. Adaptation rows ablate the cost-aware bandit, leaving an always-update policy, and the head-only action, forcing every triggered update to a full-drafter backward.

\begin{table}[t]
\centering
\caption{Ablations on Llama-3.1-8B at $T{=}0$, removing one component of the base architecture (top) or of cost-aware serving-time adaptation (bottom) per row.}
\label{tab:ablations}
\setlength{\tabcolsep}{3pt}
\resizebox{\textwidth}{!}{%
\begin{tabular}{l cccccccccccc cc}
\toprule
& \multicolumn{2}{c}{HumanEval} & \multicolumn{2}{c}{MATH-500} & \multicolumn{2}{c}{Alpaca} & \multicolumn{2}{c}{NQ} & \multicolumn{2}{c}{MT-Bench} & \multicolumn{2}{c}{WMT-23} & \multicolumn{2}{c}{Mean} \\
\cmidrule(lr){2-3} \cmidrule(lr){4-5} \cmidrule(lr){6-7} \cmidrule(lr){8-9} \cmidrule(lr){10-11} \cmidrule(lr){12-13} \cmidrule(lr){14-15}
Variant & Spd & $\tau$ & Spd & $\tau$ & Spd & $\tau$ & Spd & $\tau$ & Spd & $\tau$ & Spd & $\tau$ & Spd & $\tau$ \\
\midrule
\multicolumn{15}{c}{\textit{Base architecture}} \\
\midrule
\cellcolor{cyan!8}SpecBlock & \cellcolor{cyan!8}$3.92\times$ & \cellcolor{cyan!8}5.16 & \cellcolor{cyan!8}$3.07\times$ & \cellcolor{cyan!8}4.03 & \cellcolor{cyan!8}$3.40\times$ & \cellcolor{cyan!8}4.66 & \cellcolor{cyan!8}$3.00\times$ & \cellcolor{cyan!8}4.40 & \cellcolor{cyan!8}$3.10\times$ & \cellcolor{cyan!8}4.46 & \cellcolor{cyan!8}$2.79\times$ & \cellcolor{cyan!8}3.75 & \cellcolor{cyan!8}$3.21\times$ & \cellcolor{cyan!8}4.41 \\
$-$ prefix broadcast               & $3.69\times$ & 4.96 & $2.95\times$ & 3.73 & $3.08\times$ & 4.32 & $2.84\times$ & 4.22 & $2.83\times$ & 4.08 & $2.57\times$ & 3.49 & $2.99\times$ & 4.13 \\
$-$ layer-wise shift               & $3.65\times$ & 4.85 & $2.79\times$ & 3.55 & $3.01\times$ & 4.13 & $2.87\times$ & 4.13 & $2.78\times$ & 3.97 & $2.62\times$ & 3.49 & $2.95\times$ & 4.02 \\
$-$ valid-prefix curriculum        & $3.76\times$ & 5.06 & $2.94\times$ & 3.94 & $3.27\times$ & 4.53 & $2.86\times$ & 4.15 & $2.87\times$ & 4.15 & $2.69\times$ & 3.56 & $3.06\times$ & 4.23 \\
$-$ rank-guided branching          & $3.80\times$ & 4.84 & $2.96\times$ & 3.87 & $3.40\times$ & 4.68 & $2.92\times$ & 4.32 & $3.00\times$ & 4.26 & $2.69\times$ & 3.54 & $3.13\times$ & 4.25 \\
\midrule
\multicolumn{15}{c}{\textit{Cost-aware serving-time adaptation}} \\
\midrule
\cellcolor{cyan!8}SpecBlock+adapt & \cellcolor{cyan!8}--- & \cellcolor{cyan!8}--- & \cellcolor{cyan!8}$3.14\times$ & \cellcolor{cyan!8}4.23 & \cellcolor{cyan!8}$3.47\times$ & \cellcolor{cyan!8}4.72 & \cellcolor{cyan!8}$3.51\times$ & \cellcolor{cyan!8}5.41 & \cellcolor{cyan!8}--- & \cellcolor{cyan!8}--- & \cellcolor{cyan!8}$2.81\times$ & \cellcolor{cyan!8}3.81 & \cellcolor{cyan!8}$3.24\times$ & \cellcolor{cyan!8}4.54 \\
$-$ cost-aware bandit                & --- & --- & $2.96\times$ & 4.35 & $3.41\times$ & 4.81 & $3.48\times$ & 5.26 & --- & --- & $2.74\times$ & 3.75 & $3.14\times$ & 4.54 \\
$-$ head-only action               & --- & --- & $2.91\times$ & 4.25 & $3.20\times$ & 4.97 & $3.31\times$ & 5.46 & --- & --- & $2.60\times$ & 3.65 & $3.00\times$ & 4.58 \\
\bottomrule
\end{tabular}%
}
\end{table}
\vspace{-0.3cm}

Layer-wise shift contributes the largest single-component gain in mean speedup: removing it drops Spd from $3.21\times$ to $2.95\times$ and $\tau$ from $4.41$ to $4.02$, since cross-position causal attention alone fails to retain within-block dependence at deeper positions. The drop is heaviest on long-form benchmarks where deeper-position acceptance determines chain length, with Alpaca losing $0.39\times$ and MT-Bench $0.32\times$, while shorter-answer NQ loses only $0.13\times$. Removing the prefix broadcast costs $0.22\times$ as the position queries alone cannot anchor the $K$ positions to the verified prefix. The valid-prefix curriculum and rank-guided branching add $0.15\times$ and $0.08\times$ respectively by preventing wrong-prefix supervision and reallocating verifier budget to uncertain positions.

On the adaptation side, removing the cost-aware bandit drops mean Spd by $0.10\times$ while leaving $\tau$ unchanged at $4.54$, since always-update reaches the same accepted length but pays a backward on every query. Removing the head-only action drops mean Spd by $0.24\times$ with $\tau$ slightly higher at $4.58$, as forcing every triggered update to a full backward gains only $0.04$ in $\tau$ at the cost of an order-of-magnitude longer backward. The cost-aware bandit and head-only action therefore keep adaptation cheap by skipping weak signals and routing most updates to a backward over only the lm\_head and rank head.

\section{Conclusion}

In this paper, we introduce SpecBlock, a block-iterative drafter that produces $K$ dependent positions per forward and extends the path through repeated block expansions starting from earlier positions' hidden states. A layer-wise shift carries each previous position's hidden state into every decoder layer to preserve within-block dependence, a co-trained rank head sets per-position branching to allocate verifier budget where the drafter is uncertain, and a valid-prefix curriculum masks the loss after the first wrong prediction in the block to prevent wrong prefixes from interfering with training. Beyond static drafting, a cost-aware bandit at deployment uses free verifier feedback to update the drafter selectively, only when the expected throughput gain exceeds the update cost. SpecBlock improves mean speedup over EAGLE-3 by $8$--$13\%$ across three target models, with cost-aware adaptation extending the gain to $11$--$19\%$.

\bibliographystyle{unsrt}
\bibliography{references}


\clearpage
\appendix

\section{Implementation Details}
\label{app:impl_details}

\subsection{Drafter setup and training configuration}
\label{app:training_config}

The drafter follows the target's decoder-layer design and operates at the
same hidden dimension. The decoder and LM head are trained from scratch,
while the token embedding is initialised from the target and frozen during
training. Following EAGLE-3, the drafter operates over a reduced vocabulary
of the $32{,}000$ most frequent tokens, accounting for $98.7\%$ of training
tokens.

We use AdamW with PyTorch defaults $(\beta_1, \beta_2){=}(0.9, 0.999)$ and a
global batch size of $96$. The cosine schedule warms up linearly over the
first $1.5\%$ of total update steps. Training takes approximately
$3{,}000$ A100-80GB GPU-hours per drafter. Inference runs on a single
A100-80GB GPU by default, and cost-aware adaptation additionally has a
dual-GPU variant.

\subsection{Attention Mask for SpecBlock}
\label{app:architecture}

Figure~\ref{fig:attention_mask} illustrates the attention pattern across one
cross-block iteration, with $B_{i,k}$ denoting the $k$-th draft position of
block $i$. Each drafter forward attends to three sources of context. The verified prefix contributes the $K$ keys and values that every
previously committed position produced in its own draft forward, all fully
visible to the current forward. Preceding blocks within the current verifier
iteration are visible only along the branch path. When a new block starts
from position $j$ of a preceding block, positions $0, \dots, j$ are on-path
and visible while later positions are off-path and masked. Within the
current block, attention is causal so position $k$ attends only to
positions $\le k$.

%
%

\definecolor{cPrefix}{RGB}{60,90,140}
\definecolor{cPath}{RGB}{175,95,100}
\definecolor{cCausal}{RGB}{205,180,90}

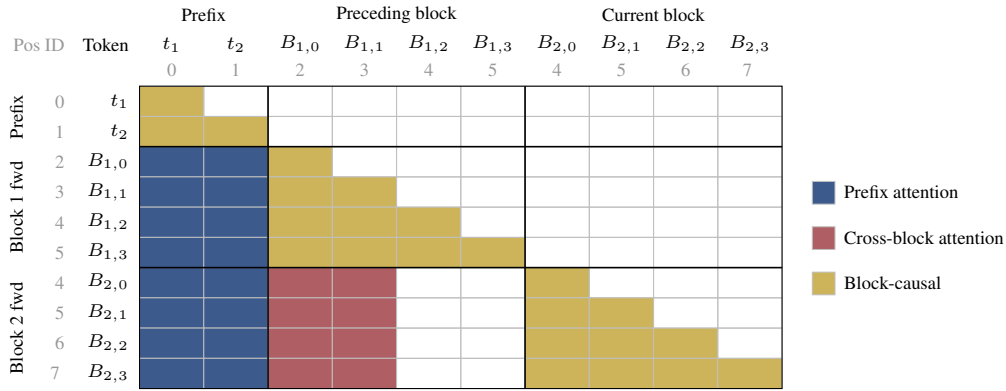
\begin{figure}[b]
  \centering
  \begin{tikzpicture}[
    cell/.style={draw=gray!50, line width=0.2pt, anchor=south west, inner sep=0pt},
    tok/.style={font=\scriptsize, align=center, inner sep=1pt},
    posid/.style={font=\scriptsize, align=center, color=gray!80, inner sep=1pt},
    region/.style={font=\scriptsize, align=center, inner sep=1pt},
    legend/.style={font=\scriptsize, anchor=west, inner sep=1pt},
  ]

    \def\cs{0.4}     
    \def\cw{0.85}    
    \pgfmathsetmacro{\xPrefEnd}{2*\cw}
    \pgfmathsetmacro{\xPrecEnd}{6*\cw}
    \pgfmathsetmacro{\xCurEnd}{10*\cw}
    \pgfmathsetmacro{\yPrefEnd}{-2*\cs}
    \pgfmathsetmacro{\yPrecEnd}{-6*\cs}
    \pgfmathsetmacro{\yBot}{-10*\cs}
    \pgfmathsetmacro{\xMidPref}{\xPrefEnd/2}
    \pgfmathsetmacro{\xMidPrec}{(\xPrefEnd+\xPrecEnd)/2}
    \pgfmathsetmacro{\xMidCur}{(\xPrecEnd+\xCurEnd)/2}
    \pgfmathsetmacro{\yMidPrefRow}{(0+\yPrefEnd)/2}
    \pgfmathsetmacro{\yMidPrecRow}{(\yPrefEnd+\yPrecEnd)/2}
    \pgfmathsetmacro{\yMidCurRow}{(\yPrecEnd+\yBot)/2}

    \foreach \q [evaluate=\q as \maxk using int(0+\q)] in {0,1} {
      \foreach \k in {0,1} {
        \pgfmathparse{\k > \maxk ? 1 : 0}
        \ifnum\pgfmathresult=1
          \draw[cell] (\k*\cw, -\q*\cs - \cs) rectangle ++(\cw, \cs);
        \else
          \filldraw[cell, fill=cCausal] (\k*\cw, -\q*\cs - \cs) rectangle ++(\cw, \cs);
        \fi
      }
      \foreach \k in {2,3,4,5,6,7,8,9} {
        \draw[cell] (\k*\cw, -\q*\cs - \cs) rectangle ++(\cw, \cs);
      }
    }

    \foreach \q in {2,3,4,5} {
      \foreach \k in {0,1} {
        \filldraw[cell, fill=cPrefix] (\k*\cw, -\q*\cs - \cs) rectangle ++(\cw, \cs);
      }
    }
    \foreach \q [evaluate=\q as \maxk using int(2 + \q - 2)] in {2,3,4,5} {
      \foreach \k in {2,3,4,5} {
        \pgfmathparse{\k > \maxk ? 1 : 0}
        \ifnum\pgfmathresult=1
          \draw[cell] (\k*\cw, -\q*\cs - \cs) rectangle ++(\cw, \cs);
        \else
          \filldraw[cell, fill=cCausal] (\k*\cw, -\q*\cs - \cs) rectangle ++(\cw, \cs);
        \fi
      }
    }
    \foreach \q in {2,3,4,5} {
      \foreach \k in {6,7,8,9} {
        \draw[cell] (\k*\cw, -\q*\cs - \cs) rectangle ++(\cw, \cs);
      }
    }

    \foreach \q in {6,7,8,9} {
      \foreach \k in {0,1} {
        \filldraw[cell, fill=cPrefix] (\k*\cw, -\q*\cs - \cs) rectangle ++(\cw, \cs);
      }
    }
    \foreach \q in {6,7,8,9} {
      \foreach \k in {2,3} {
        \filldraw[cell, fill=cPath] (\k*\cw, -\q*\cs - \cs) rectangle ++(\cw, \cs);
      }
      \foreach \k in {4,5} {
        \draw[cell] (\k*\cw, -\q*\cs - \cs) rectangle ++(\cw, \cs);
      }
    }
    \foreach \q [evaluate=\q as \maxk using int(6 + \q - 6)] in {6,7,8,9} {
      \foreach \k in {6,7,8,9} {
        \pgfmathparse{\k > \maxk ? 1 : 0}
        \ifnum\pgfmathresult=1
          \draw[cell] (\k*\cw, -\q*\cs - \cs) rectangle ++(\cw, \cs);
        \else
          \filldraw[cell, fill=cCausal] (\k*\cw, -\q*\cs - \cs) rectangle ++(\cw, \cs);
        \fi
      }
    }

    \draw[line width=0.6pt] (\xPrefEnd, 0) -- (\xPrefEnd, \yBot);
    \draw[line width=0.6pt] (\xPrecEnd, 0) -- (\xPrecEnd, \yBot);
    \draw[line width=0.6pt] (0, \yPrefEnd) -- (\xCurEnd, \yPrefEnd);
    \draw[line width=0.6pt] (0, \yPrecEnd) -- (\xCurEnd, \yPrecEnd);
    \draw[line width=0.4pt] (0, 0) rectangle (\xCurEnd, \yBot);

    \foreach \k/\name/\pid in {
      0/{$t_1$}/0, 1/{$t_2$}/1,
      2/{$B_{1,0}$}/2, 3/{$B_{1,1}$}/3, 4/{$B_{1,2}$}/4, 5/{$B_{1,3}$}/5,
      6/{$B_{2,0}$}/4, 7/{$B_{2,1}$}/5, 8/{$B_{2,2}$}/6, 9/{$B_{2,3}$}/7
    } {
      \node[tok] at (\k*\cw + \cw/2, 0.55) {\name};
      \node[posid] at (\k*\cw + \cw/2, 0.22) {\pid};
    }

    \node[region] at (\xMidPref, 0.95) {Prefix};
    \node[region] at (\xMidPrec, 0.95) {Preceding block};
    \node[region] at (\xMidCur, 0.95) {Current block};

    \foreach \q/\name/\pid in {
      0/{$t_1$}/0, 1/{$t_2$}/1,
      2/{$B_{1,0}$}/2, 3/{$B_{1,1}$}/3, 4/{$B_{1,2}$}/4, 5/{$B_{1,3}$}/5,
      6/{$B_{2,0}$}/4, 7/{$B_{2,1}$}/5, 8/{$B_{2,2}$}/6, 9/{$B_{2,3}$}/7
    } {
      \node[tok, anchor=east] at (-0.10, -\q*\cs - \cs/2) {\name};
      \node[posid, anchor=east] at (-0.95, -\q*\cs - \cs/2) {\pid};
    }
    \node[tok, anchor=east] at (-0.10, 0.55) {Token};
    \node[posid, anchor=east] at (-0.95, 0.55) {Pos ID};

    \node[region, rotate=90, anchor=south] at (-1.50, \yMidPrefRow) {Prefix};
    \node[region, rotate=90, anchor=south] at (-1.50, \yMidPrecRow) {Block 1 fwd};
    \node[region, rotate=90, anchor=south] at (-1.50, \yMidCurRow) {Block 2 fwd};

    \pgfmathsetmacro{\xLeg}{\xCurEnd + 0.4}
    \pgfmathsetmacro{\yLegA}{-1.40}
    \pgfmathsetmacro{\yLegB}{-2.00}
    \pgfmathsetmacro{\yLegC}{-2.60}
    \filldraw[fill=cPrefix, draw=gray!50, line width=0.2pt]
      (\xLeg, \yLegA - 0.16) rectangle ++(0.30, 0.30);
    \node[legend] at (\xLeg + 0.38, \yLegA) {Prefix attention};
    \filldraw[fill=cPath, draw=gray!50, line width=0.2pt]
      (\xLeg, \yLegB - 0.16) rectangle ++(0.30, 0.30);
    \node[legend] at (\xLeg + 0.38, \yLegB) {Cross-block attention};
    \filldraw[fill=cCausal, draw=gray!50, line width=0.2pt]
      (\xLeg, \yLegC - 0.16) rectangle ++(0.30, 0.30);
    \node[legend] at (\xLeg + 0.38, \yLegC) {Block-causal};

  \end{tikzpicture}
  \caption{Attention pattern across one cross-block iteration with $K{=}4$.
    Block~2 branches from $B_{1,1}$, so the path leading to block~2 consists
    of $B_{1,0}$ and $B_{1,1}$. The prefix's tokens attend to themselves
    causally. Each block forward sees the verified prefix as Prefix attention
    plus itself as Block-causal. Block~2's queries additionally attend to
    the on-path positions $B_{1,0}, B_{1,1}$ as Cross-block attention, while
    the off-path positions $B_{1,2}, B_{1,3}$ are masked.}
  \label{fig:attention_mask}
\end{figure}

\subsection{Cost-aware adaptation}
\label{app:cost_aware_adapt}

\paragraph{Bandit lifecycle.}
A short warmup of $10$ queries collects the baseline throughput before any
training event. The bandit then enters a cold-start phase of $8$ events
with $\varepsilon$-greedy exploration, with $\varepsilon$ decaying linearly
from $0.30$ to $0.10$, so the value estimates $v_{\text{head}}$ and
$v_{\text{full}}$ are seeded from each action. After each train event, the
next $2$ queries are blocked from triggering further updates, and the EWMA
update on the previous mode's reward is skipped when its trigger signal
$s_{\text{trig}} < 5$ to avoid amplifying noise from low-signal updates.


\begin{figure}[t]
  \centering
  \includegraphics[width=\linewidth]{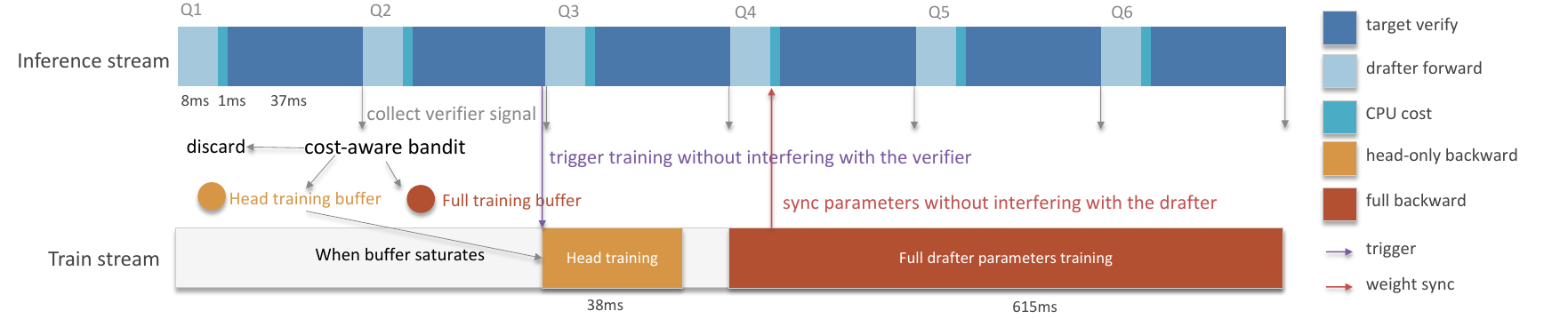}
  \caption{Cost-aware adaptation scheduling. The cost-aware bandit ingests
    the verifier signal at every query end and routes it into the head or
    full training buffer. When a buffer saturates, the corresponding
    backward fires on the train stream and weights sync back, both timed
    to avoid interfering with the verifier and the drafter. The schedule
    applies to single-GPU and dual-GPU deployments.}
  \label{fig:adapt_timing}
\end{figure}

\paragraph{Scheduling and synchronisation.}
Figure~\ref{fig:adapt_timing} illustrates the per-query schedule. Training
fires on the train stream only when its buffer saturates, and is timed so
the backward does not interfere with the verifier. Weight synchronisation
between $\theta_{\text{train}}$ and $\theta_{\text{inf}}$ is timed during
target-verify windows when the drafter is otherwise idle. In single-GPU
deployment the train stream runs on the same device as inference but uses
a separate CUDA stream, so its kernels share the GPU's streaming
multiprocessors with inference rather than preempting it. In dual-GPU
deployment the train stream lives on a second device and the two
drafter copies sync via periodic in-place weight copy.

\paragraph{Per-query control flow.}
Algorithm~\ref{alg:adapt} ties together the bandit, the train stream,
and the drift control. The verifier-derived signal predicts the net gain
of each non-skip action at query time, and its value $s_{\text{trig}}$ at
the query that fires an update later gates the EWMA update of that
update's reward. When a buffer saturates, the corresponding
backward fires on the train stream, and the trained weights are synced
into $\theta_{\text{inf}}$ during the next target-verify window. A
monotonic drop in accepted length over three windows triggers a
rollback to the last good checkpoint.

\begin{algorithm}[tb]
\caption{Cost-aware adaptation, per query.}
\label{alg:adapt}
\begin{algorithmic}[1]
\Require pre-deployment drafter $\theta_0$, two-copy state $(\theta_{\text{inf}}, \theta_{\text{train}})$, value estimates $(v_{\text{head}}, v_{\text{full}})$, head and full buffers $(\mathcal{B}_{\text{head}}, \mathcal{B}_{\text{full}})$, EWMA decay $\alpha$, KL weight $\lambda$, signal threshold $s_{\min}$, measurement interval length $N$
\State complete drafting and verification on $\theta_{\text{inf}}$
\State $s \gets \sum_{k\,\in\,\text{rejected}} (1 - r_{t,k})$ \Comment{verifier-derived signal}
\State $\widehat{\Delta}_{\text{tp}}(\text{head}) \gets s \cdot v_{\text{head}}$;\quad $\widehat{\Delta}_{\text{tp}}(\text{full}) \gets s \cdot v_{\text{full}}$
\If{$\widehat{\Delta}_{\text{tp}}(\text{head}) \le 0$ \textbf{and} $\widehat{\Delta}_{\text{tp}}(\text{full}) \le 0$}
  \State $a \gets \text{skip}$
\Else
  \State $a \gets \arg\max_{\,\text{action} \in \{\text{head}, \text{full}\}} \widehat{\Delta}_{\text{tp}}(\text{action})$
\EndIf
\If{$a \neq \text{skip}$}
  \State push (rejected positions, target distributions, $s$) into $\mathcal{B}_a$
\EndIf
\If{$\mathcal{B}_a$ saturates}
  \State $s_{\text{trig}} \gets s$;\quad $a_{\text{trig}} \gets a$ \Comment{snapshot trigger signal and action}
  \State \textbf{on train stream:} step $\theta_{\text{train}}$ on $\mathcal{L}^{\text{adapt}}_a + \lambda\,\mathrm{KL}(p_{\theta_{\text{train}}} \,\|\, p_{\theta_0})$
  \State clear $\mathcal{B}_a$
  \State during the next target-verify window: $\theta_{\text{inf}} \gets \theta_{\text{train}}$
  \State open measurement interval of $N$ queries to record $\Delta_{\text{tp}}^{\text{observed}}$
\EndIf
\If{measurement interval closes \textbf{and} $s_{\text{trig}} \ge s_{\min}$}
  \State $v_{a_{\text{trig}}} \gets (1 - \alpha)\,v_{a_{\text{trig}}} + \alpha\,\Delta_{\text{tp}}^{\text{observed}}/s_{\text{trig}}$ \Comment{EWMA update}
\EndIf
\If{accepted length decreases monotonically over the last $3$ windows}
  \State revert $\theta_{\text{inf}}$ to the last good checkpoint, reset $v_{\text{head}}, v_{\text{full}}$
\EndIf
\end{algorithmic}
\end{algorithm}

\subsection{Baselines and benchmarks}
\label{app:baselines_benchmarks}

\paragraph{Baselines.}
We compare against six drafting baselines.
\begin{itemize}
  \item \textbf{Standard speculative sampling
        (SpS)}~\citep{leviathan2023fast,chen2023accelerating} samples a
        chain of future tokens autoregressively from a smaller
        off-the-shelf drafter and lets the target verify the chain in
        one parallel forward. We pair Llama-3.1-8B with Llama-3.2-1B,
        and the Qwen3 targets with Qwen3-0.6B.
  \item \textbf{Medusa}~\citep{cai2024medusa} attaches $K$ independent
        decoding heads at fixed offsets to the target's last hidden
        state, with no cross-position attention or layer-wise
        dependence.
  \item \textbf{ParallelSpec}~\citep{xiao2024parallelspec} predicts $K$
        future tokens in one drafter forward by appending $K$ learnable
        \texttt{[MASK]} tokens after the prefix and reading their
        last-layer hidden states, under a group-wise causal mask that
        blocks attention from \texttt{[MASK]}s in earlier parallel
        groups.
  \item \textbf{Falcon}~\citep{hu2024falcon} is the closest blockwise
        prior work. It drafts semi-autoregressive blocks of $k$ tokens
        with a hybrid drafter combining LSTM layers with
        relaxed-causal-mask self-attention, letting positions inside
        the same $k{\times}k$ block attend to one another, and verifies
        through a custom-designed static decoding tree.
  \item \textbf{EAGLE-3}~\citep{li2025eagle3} is an autoregressive
        token-level drafter that fuses the target's low-, mid-, and
        high-level hidden states as input, replacing the top-layer-only
        reuse of EAGLE-1/2~\citep{li2024eagle,li2024eagle2}, and grows
        a dynamic draft tree depth by depth by calling the drafter
        once per added depth.
  \item \textbf{SpecBlock+OSD} instantiates Online Speculative
        Decoding~\citep{liu2024online} on the SpecBlock drafter as the
        always-update baseline. Rejection-position pairs of the draft
        and target distributions are logged to a replay buffer, and the
        drafter is updated every eight queries via forward-KL
        distillation, with no bandit gating over which signals to
        apply.
\end{itemize}

\paragraph{Benchmarks.}
The six benchmarks vary substantially in prompt count, from $80$ multi-turn
dialogues in MT-Bench and $164$ code prompts in HumanEval, through $500$
competition problems in MATH-500 and $549$ translation pairs in WMT-23, up
to $4{,}000$ instruction prompts in Alpaca and $3{,}610$ open-domain
questions in Natural Questions. Each prompt is rendered through the
target's official chat template, with thinking mode disabled for Qwen3 so
the verifier output is comparable to Llama, and passed once through the
speculative decoding loop under greedy decoding ($T{=}0$) or stochastic
decoding ($T{=}1.0$). Generation is capped at $1{,}024$ newly committed
tokens. We report speedup as the wall-clock ratio over vanilla
autoregressive decoding on the same prompt set, accepted length $\tau$ as
the average length committed per verifier call, and drafting cost
$T_\mathcal{D}\%$ as the share of per-iteration latency spent on drafter
forwards. Cost-aware adaptation is reported only on MATH-500, Alpaca, NQ,
and WMT-23, whose streaming-prompt counts are large enough for the
adaptation backward to amortize its cost; HumanEval and MT-Bench leave too
little traffic for the bandit's value estimates to converge.

\subsection{Inference procedure}
\label{app:inference_procedure}

Algorithm~\ref{alg:inference} traces one verifier iteration of SpecBlock.
The drafter is invoked at most $M$ times to grow a tree of up to $M{\cdot}K$
depth, the verifier scores all candidates in a single parallel forward, and
the longest accepted prefix is committed before the next iteration starts.
The first block conditions on the target's multi-layer features at the
prefix-end position, while later blocks condition on the drafter's own
cached last-layer state at the starting position, bypassing the projection
$W_{\text{cond}}$. Per-position branching width follows the rank head's
bucket through the map $b(\cdot)$. We recommend two configurations of
$b(\cdot)$ over the four buckets defined in \S\ref{sec:method:rank}:
$[2, 4, 10, 0]$, which concentrates tree budget on the less confident
buckets where the target token sits deeper and gives up on the rank$>$10 bucket, and $[2, 4, 6, 4]$, which
trims the wide rank-$5$--$10$ bucket and keeps a $4$-candidate fallback at
give-up positions for out-of-distribution traffic.

\begin{algorithm}[tb]
\caption{SpecBlock inference, one verifier iteration.}
\label{alg:inference}
\begin{algorithmic}[1]
\Require target $\mathcal{M}$, drafter $\mathcal{D}_\theta$, last verified token $x_t$ and multi-layer features $(h^{\text{low}}_t, h^{\text{mid}}_t, h^{\text{top}}_t)$ from $\mathcal{M}$'s most recent forward, block depth $K$, block budget $M$, bucket-to-branching map $b(\cdot)$
\Ensure tokens committed in this iteration
\State $c_t \gets W_{\text{cond}}\,[h^{\text{low}}_t, h^{\text{mid}}_t, h^{\text{top}}_t]$
\State $\mathrm{starts} \gets \{(c_t,\, \mathrm{embed}(x_t))\}$ \Comment{first-block starting point uses target features}
\State $\mathrm{tree} \gets $ empty draft tree rooted at $x_t$
\For{$m = 1$ \textbf{to} $M$}
  \State $\{p_{i,k},\, h^{(L)}_{i,k},\, \mathrm{bkt}_{i,k}\}_{i,\,k=1}^{K} \gets \mathcal{D}_\theta(\mathrm{starts})$ \Comment{batched forward, $K$ positions per starting point}
  \For{each starting point $i$ and each position $k$}
    \State attach the top-$b(\mathrm{bkt}_{i,k})$ tokens of $p_{i,k}$ to $\mathrm{tree}$ as candidates at position $(i, k)$
  \EndFor
  \State $\mathrm{starts} \gets$ next-block starting points selected by the rank head, each carrying its cached $h^{(L)}_{i,k}$ and sampled token
\EndFor
\State $(\mathrm{prefix},\, \mathrm{bonus}) \gets \mathcal{M}.\mathrm{verify}(\mathrm{tree})$ \Comment{single parallel target forward}
\State \Return $\mathrm{prefix} \,\cup\, \mathrm{bonus}$
\end{algorithmic}
\end{algorithm}

\section{Rank head}
\label{app:rank_head}

\paragraph{Distribution summary features.}
Section~\ref{sec:method:rank} states that the rank head reads a
$15$-dimensional summary $\psi(p_{t,k})$ of the draft distribution
alongside the hidden state. We detail those $15$ dimensions here. The
summary captures how peaky or flat the distribution is through three
kinds of signal. The bulk of $\psi$ records the log-probability profile
of the top-$10$ tokens, which describes how mass spreads across the
most likely candidates and contributes ten dimensions. Three further
dimensions capture the logit gaps between the top token and its
rank-$2$, rank-$3$, and rank-$5$ competitors, telling the rank head how
distinguishable the leader is from close runners-up. The remaining two
dimensions are scalar summaries: the probability of the top token, and
the entropy of the distribution.

\paragraph{Classification quality.}
\label{app:rank_quality}
The rank head is supervised as a four-way classifier over the bucket
labels defined in \S\ref{sec:method:rank}, and its classification quality
directly affects how the verifier budget is spent. We evaluate it as a
standalone classifier on $\sim 72{,}000$ validation positions of the
SpecBlock drafter for Llama-3.1-8B, restricted to positions whose
valid-prefix mask is one. At each such position we record the predicted
bucket and the ground-truth bucket derived from the target token's rank
within $p_{t,k}$.

Table~\ref{tab:rank_quality} reports per-bucket precision, recall, and
F1 together with class frequency. The class frequencies are sharply
imbalanced, with $b_0$ accounting for $75.2\%$ of positions while $b_2$
and $b_3$ each fall under $6\%$. The two extreme buckets are the
easiest to classify. The confident bucket $b_0$ reaches precision
$0.978$ since the drafter is well-calibrated when the target sits at
rank $1$, and the give-up bucket $b_3$ reaches F1 $0.822$ since
high-rank positions carry clear distribution-shape signals such as flat
or multi-modal $p_{t,k}$. The two middle buckets $b_1$ and $b_2$ are
the hardest, with F1 around $0.46$--$0.50$, and most errors fall
between the two adjacent buckets.

\begin{table}[tb]
\centering
\caption{Rank head classification quality on $\sim 72{,}000$ held-out positions of the SpecBlock drafter for Llama-3.1-8B.}
\label{tab:rank_quality}
\begin{tabular}{lcccc}
\toprule
Bucket & Frequency (\%) & Precision & Recall & F1 \\
\midrule
$b_0$ (rank ${=}1$)        & 75.2 & 0.978 & 0.687 & 0.807 \\
$b_1$ (rank $\in [2, 4]$)  & 15.0 & 0.528 & 0.476 & 0.501 \\
$b_2$ (rank $\in [5, 10]$) &  4.1 & 0.468 & 0.448 & 0.458 \\
$b_3$ (rank $>10$)         &  5.7 & 0.854 & 0.792 & 0.822 \\
\bottomrule
\end{tabular}
\end{table}

\section{Per-position acceptance rate}
\label{app:per_position_alpha}

We measure two acceptance-rate diagnostics averaged across benchmarks.
$\alpha_k$ is the probability that the drafter's greedy token at chain
position $k$ matches the target's greedy continuation, assuming
positions $1,\ldots,k{-}1$ have all already matched, swept over
$k=1,\ldots,K{\cdot}M{=}8$ across two cross-block iterations. The chain
position $k$ does not correspond to a fixed block-internal index, since
the boundary depends on how many positions of block~0 are taken before
block~1 starts. $\alpha_{m,j}$ replots the rate at position $j$ of
block $m$ under the same prior-match filter, with block~1 additionally
assuming block~0 was fully accepted.

\definecolor{llamablue}{RGB}{31,119,180}
\definecolor{qwenorange}{RGB}{255,127,14}
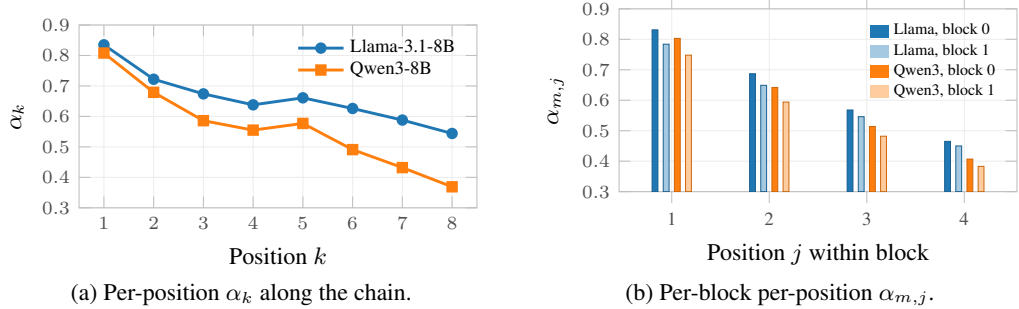
\begin{figure}[t]
\centering
\begin{minipage}[t]{0.49\linewidth}
\centering
\begin{tikzpicture}
\begin{axis}[
  width=\linewidth, height=4cm,
  xlabel={Position $k$}, ylabel={$\alpha_k$},
  xmin=0.5, xmax=8.5, ymin=0.3, ymax=0.9,
  xtick={1,2,3,4,5,6,7,8},
  ytick={0.3,0.4,0.5,0.6,0.7,0.8,0.9},
  legend style={font=\scriptsize, draw=none, fill=none, anchor=north east, at={(0.98,0.98)}, row sep=-2pt},
  legend cell align=left,
  axis line style={gray!50},
  tick style={gray!50},
  grid=major, grid style={gray!15, line width=0.3pt},
  tick label style={font=\scriptsize, color=black!70},
  label style={font=\small},
]
\addplot[llamablue, line width=1.1pt, mark=*, mark size=1.6pt] coordinates {(1,0.835)(2,0.722)(3,0.674)(4,0.638)(5,0.661)(6,0.626)(7,0.588)(8,0.544)};
\addlegendentry{Llama-3.1-8B}
\addplot[qwenorange, line width=1.1pt, mark=square*, mark size=1.6pt] coordinates {(1,0.808)(2,0.679)(3,0.586)(4,0.555)(5,0.577)(6,0.491)(7,0.432)(8,0.369)};
\addlegendentry{Qwen3-8B}
\end{axis}
\end{tikzpicture}\\
\small (a) Per-position $\alpha_k$ along the chain.
\end{minipage}\hfill
\begin{minipage}[t]{0.49\linewidth}
\centering
\begin{tikzpicture}
\begin{axis}[
  width=\linewidth, height=4cm,
  ybar, bar width=2.2pt,
  enlarge x limits=0.18,
  xlabel={Position $j$ within block}, ylabel={$\alpha_{m,j}$},
  symbolic x coords={1,2,3,4},
  xtick=data,
  ymin=0.3, ymax=0.9,
  ytick={0.3,0.4,0.5,0.6,0.7,0.8,0.9},
  legend style={font=\tiny, draw=none, fill=none, anchor=north east, at={(0.98,0.98)}, row sep=-2pt},
  legend cell align=left,
  legend image code/.code={\draw[#1, draw=none] (0cm,-0.08cm) rectangle (0.25cm,0.08cm);},
  axis line style={gray!50},
  tick style={gray!50},
  grid=major, grid style={gray!15, line width=0.3pt},
  tick label style={font=\scriptsize, color=black!70},
  label style={font=\small},
]
\addplot[fill=llamablue, draw=llamablue!80!black, line width=0.3pt] coordinates {(1,0.831)(2,0.687)(3,0.568)(4,0.465)};
\addlegendentry{Llama, block 0}
\addplot[fill=llamablue!40, draw=llamablue!80!black, line width=0.3pt] coordinates {(1,0.784)(2,0.649)(3,0.546)(4,0.450)};
\addlegendentry{Llama, block 1}
\addplot[fill=qwenorange, draw=qwenorange!80!black, line width=0.3pt] coordinates {(1,0.803)(2,0.642)(3,0.514)(4,0.407)};
\addlegendentry{Qwen3, block 0}
\addplot[fill=qwenorange!40, draw=qwenorange!80!black, line width=0.3pt] coordinates {(1,0.748)(2,0.594)(3,0.482)(4,0.383)};
\addlegendentry{Qwen3, block 1}
\end{axis}
\end{tikzpicture}\\
\small (b) Per-block per-position $\alpha_{m,j}$.
\end{minipage}
\caption{Acceptance rate diagnostics averaged across benchmarks. (a) Per-position $\alpha_k$ along the chain to depth $K{\cdot}M{=}8$, computed under the assumption that positions $1,\ldots,k-1$ all match the target's greedy continuation. (b) Per-block per-position $\alpha_{m,j}$ at position $j$ of block $m$, computed under the assumption that positions $1,\ldots,j-1$ of the same block match; block~1 additionally assumes block~0 was fully accepted, since block~1 starts from the drafter's own cached $h^{(L)}$ at the last position of block~0.}
\label{fig:acceptance_diagnostics}
\end{figure}

Figure~\ref{fig:acceptance_diagnostics}(a) shows a smooth decay on both
targets, with $\alpha_1$ above $0.80$ and $\alpha_8$ around
$0.37$--$0.54$. Qwen3-8B drops sharper than Llama-3.1-8B, ending at
$\alpha_8{=}0.369$ while Llama-3.1-8B retains $0.544$.
Figure~\ref{fig:acceptance_diagnostics}(b) shows that $\alpha_{m,j}$
decays monotonically across the four positions within each block.
Position~1 of block~1 reaches $0.784$ on Llama-3.1-8B and $0.748$ on
Qwen3-8B, both well above the last position of block~0 at $0.465$ and
$0.407$, even though block~1 starts from the drafter's own cached state
rather than the target's. The block boundary therefore acts as a
recovery mechanism, supporting the block-iterative design over a single
longer block of length $K{\cdot}M$.

\section{Adaptation deployment}
\label{app:adaptation_experiments}

\subsection{Single-GPU and dual-GPU deployment}
\label{app:single_vs_dual}

Cost-aware adaptation supports two deployment regimes: single-GPU, where
the train stream runs on the same device as inference but on a separate
CUDA stream, and dual-GPU, where the train stream runs on a second
device and the two drafter copies sync via periodic in-place weight
copy. Table~\ref{tab:single_vs_dual} compares the two regimes on the
three target models at $T{=}0$ and $T{=}1$, reporting raw Spd over
vanilla decoding and accepted length $\tau$ for SpecBlock without
adaptation, SpecBlock+adapt single-GPU, and SpecBlock+adapt dual-GPU.

\begin{table}[t]
\centering
\caption{SpecBlock+adapt under single-GPU and dual-GPU deployment across three target models. Subscripts on each cell show the absolute gain over SpecBlock without adaptation. Spd is the speedup over vanilla decoding and $\tau$ is the average accepted length per verifier call.}
\label{tab:single_vs_dual}
\setlength{\tabcolsep}{4pt}
\resizebox{\textwidth}{!}{%
\begin{tabular}{ll cc cc cc cc}
\toprule
& & \multicolumn{4}{c}{$T{=}0$} & \multicolumn{4}{c}{$T{=}1$} \\
\cmidrule(lr){3-6} \cmidrule(lr){7-10}
& & \multicolumn{2}{c}{Spd} & \multicolumn{2}{c}{$\tau$} & \multicolumn{2}{c}{Spd} & \multicolumn{2}{c}{$\tau$} \\
\cmidrule(lr){3-4} \cmidrule(lr){5-6} \cmidrule(lr){7-8} \cmidrule(lr){9-10}
Model & Bench & single & dual & single & dual & single & dual & single & dual \\
\midrule
\multirow{4}{*}{Llama-3.1-8B} & MATH-500 & $3.14_{+0.07}$ & $3.12_{+0.05}$ & $4.23_{+0.20}$ & $4.20_{+0.17}$ & $1.78_{+0.04}$ & $1.77_{+0.03}$ & $3.31_{+0.07}$ & $3.26_{+0.02}$ \\
& WMT-23 & $2.81_{+0.02}$ & $2.93_{+0.14}$ & $3.81_{+0.06}$ & $3.83_{+0.08}$ & $2.36_{+0.05}$ & $2.35_{+0.04}$ & $3.57_{+0.16}$ & $3.57_{+0.16}$ \\
& Alpaca & $3.47_{+0.07}$ & $3.46_{+0.06}$ & $4.72_{+0.06}$ & $4.69_{+0.03}$ & $2.72_{+0.15}$ & $2.66_{+0.09}$ & $4.22_{+0.20}$ & $4.19_{+0.17}$ \\
& NQ & $3.51_{+0.51}$ & $3.37_{+0.37}$ & $5.41_{+1.01}$ & $5.46_{+1.06}$ & $2.39_{+0.47}$ & $2.37_{+0.45}$ & $4.57_{+0.97}$ & $4.60_{+1.00}$ \\
\midrule
\multirow{4}{*}{Qwen3-8B} & MATH-500 & $2.61_{+0.08}$ & $2.59_{+0.06}$ & $3.90_{+0.19}$ & $3.89_{+0.18}$ & $2.67_{+0.07}$ & $2.64_{+0.04}$ & $3.79_{+0.16}$ & $3.70_{+0.07}$ \\
& WMT-23 & $2.34_{+0.04}$ & $2.42_{+0.12}$ & $3.36_{+0.08}$ & $3.36_{+0.08}$ & $2.19_{+0.08}$ & $2.16_{+0.05}$ & $3.31_{+0.12}$ & $3.31_{+0.12}$ \\
& Alpaca & $2.64_{+0.06}$ & $2.63_{+0.05}$ & $3.78_{+0.05}$ & $3.77_{+0.04}$ & $2.19_{+0.13}$ & $2.15_{+0.09}$ & $3.69_{+0.10}$ & $3.66_{+0.07}$ \\
& NQ & $2.69_{+0.43}$ & $2.55_{+0.29}$ & $3.75_{+0.54}$ & $3.78_{+0.57}$ & $2.27_{+0.46}$ & $2.24_{+0.43}$ & $3.55_{+0.49}$ & $3.54_{+0.48}$ \\
\midrule
\multirow{4}{*}{Qwen3-32B} & MATH-500 & $2.55_{+0.07}$ & $2.53_{+0.05}$ & $3.73_{+0.20}$ & $3.71_{+0.18}$ & $2.55_{+0.07}$ & $2.51_{+0.03}$ & $3.64_{+0.16}$ & $3.56_{+0.08}$ \\
& WMT-23 & $2.24_{+0.04}$ & $2.31_{+0.11}$ & $3.22_{+0.07}$ & $3.21_{+0.06}$ & $2.14_{+0.07}$ & $2.10_{+0.03}$ & $3.22_{+0.12}$ & $3.22_{+0.12}$ \\
& Alpaca & $2.51_{+0.03}$ & $2.52_{+0.04}$ & $3.61_{+0.07}$ & $3.56_{+0.02}$ & $2.06_{+0.12}$ & $2.02_{+0.08}$ & $3.53_{+0.11}$ & $3.48_{+0.06}$ \\
& NQ & $2.57_{+0.40}$ & $2.45_{+0.28}$ & $3.61_{+0.54}$ & $3.62_{+0.55}$ & $2.19_{+0.44}$ & $2.16_{+0.41}$ & $3.40_{+0.48}$ & $3.41_{+0.49}$ \\
\bottomrule
\end{tabular}%
}
\end{table}

The two regimes deliver comparable gains on every benchmark and across
all three target models, with neither regime dominating. Single-GPU
matches or even exceeds dual-GPU on Spd for MATH-500, Alpaca, and NQ across
both temperatures, while dual-GPU wins on WMT-23 at $T{=}0$. The $\tau$
gap between the two regimes stays within $0.05$ on most benchmarks, and
both regimes recover most of the OOD acceptance loss on NQ. The
dual-GPU regime additionally pays a cross-device weight sync latency,
so frequent syncs do not automatically translate into higher throughput
than the single-GPU regime, where the train stream shares the streaming
multiprocessors with inference but avoids cross-device transfer.
Cost-aware adaptation is therefore viable on a single-GPU deployment,
and the dual-GPU variant is an option when an extra device is
available for the train stream.

\subsection{Mixed-task adaptation}
\label{app:mixed_task}

Production traffic is rarely a single homogeneous benchmark: user
requests typically mix tasks such as math, translation, instruction
following, and QA, and may revisit similar prompt patterns over time
as the same users return. To inspect adaptation behavior under such
heterogeneous and repeated traffic, we construct a mixed stream by
sampling equal proportions of prompts from MATH-500, WMT-23, Alpaca,
and NQ, with a total stream size of $2$K queries. We then sweep the
number of full passes over the mixed stream
$N \in \{1, 2, 4, 6, 8\}$ to mimic the steady-state regime where the
drafter has seen the user's task distribution multiple times.

\begin{table}[t]
\centering
\caption{SpecBlock+adapt on the $2$K mixed stream after $N$ epochs of adaptation over the stream. The first row reports SpecBlock without adaptation on the same stream as a no-adapt baseline; subscripts on each adapted row show the absolute gain over that baseline. Spd is the speedup over vanilla decoding and $\tau$ is the average accepted length per verifier call.}
\label{tab:mixed_task_2k}
\setlength{\tabcolsep}{4pt}
\resizebox{\textwidth}{!}{%
\begin{tabular}{l cc cc cc cc}
\toprule
& \multicolumn{4}{c}{Llama-3.1-8B} & \multicolumn{4}{c}{Qwen3-8B} \\
\cmidrule(lr){2-5} \cmidrule(lr){6-9}
& \multicolumn{2}{c}{Spd} & \multicolumn{2}{c}{$\tau$} & \multicolumn{2}{c}{Spd} & \multicolumn{2}{c}{$\tau$} \\
\cmidrule(lr){2-3} \cmidrule(lr){4-5} \cmidrule(lr){6-7} \cmidrule(lr){8-9}
$N$ & single & dual & single & dual & single & dual & single & dual \\
\midrule
no adapt & $3.05$ & $3.05$ & $4.19$ & $4.19$ & $2.41$ & $2.41$ & $3.47$ & $3.47$ \\
\midrule
$1$ & $3.11_{+0.06}$ & $3.11_{+0.06}$ & $4.26_{+0.07}$ & $4.28_{+0.09}$ & $2.50_{+0.09}$ & $2.51_{+0.10}$ & $3.61_{+0.14}$ & $3.63_{+0.16}$ \\
$2$ & $3.13_{+0.08}$ & $3.13_{+0.08}$ & $4.31_{+0.12}$ & $4.32_{+0.13}$ & $2.54_{+0.13}$ & $2.55_{+0.14}$ & $3.66_{+0.19}$ & $3.68_{+0.21}$ \\
$4$ & $3.14_{+0.09}$ & $3.15_{+0.10}$ & $4.34_{+0.15}$ & $4.36_{+0.17}$ & $2.58_{+0.17}$ & $2.58_{+0.17}$ & $3.68_{+0.21}$ & $3.72_{+0.25}$ \\
$6$ & $3.16_{+0.11}$ & $3.16_{+0.11}$ & $4.38_{+0.19}$ & $4.41_{+0.22}$ & $2.62_{+0.21}$ & $2.63_{+0.22}$ & $3.72_{+0.25}$ & $3.74_{+0.27}$ \\
$8$ & $3.17_{+0.12}$ & $3.17_{+0.12}$ & $4.41_{+0.22}$ & $4.45_{+0.26}$ & $2.65_{+0.24}$ & $2.65_{+0.24}$ & $3.77_{+0.30}$ & $3.76_{+0.29}$ \\
\bottomrule
\end{tabular}%
}
\end{table}

Both Spd and $\tau$ grow monotonically with the number of adapt epochs
$N$ on both target models and both deployment regimes. After $N{=}8$
epochs, the adaptation lifts $\tau$ by $0.22$--$0.30$ over the no-adapt
baseline and Spd by roughly $3.9$--$10.0\%$, with the larger gains on
Qwen3-8B where the original drafter has more headroom. Dual-GPU is
consistently a small step ahead of single-GPU but the gap stays under
$0.05$ in $\tau$ and under $1\%$ in Spd, again echoing that cross-device
sync latency offsets some of the benefit of separating the train
stream from inference. The monotonic growth confirms that cost-aware
adaptation can extract additional accepted length when the same mixed
traffic is revisited, mirroring a deployment scenario where returning
users supply repeated task patterns to the same drafter instance.

\section{Case Study}
\label{app:case_study}

To illustrate the structural difference between SpecBlock and EAGLE-3 trees and how each translates into accepted tokens, we run both drafters on the same prompt with the Qwen3-8B target under greedy decoding. Figure~\ref{fig:case_response} shows the first $30$ committed tokens of each verifier-committed response, with green shading for tokens drafted and accepted by the verifier and red shading for bonus tokens sampled from the target after acceptance. The two drafters commit nearly the same content, and most of the divergence between them is in which positions land as bonuses rather than as drafted accepts.

\definecolor{acceptbg}{RGB}{210, 240, 215}  
\definecolor{bonusbg}{RGB}{255, 205, 205}   

\begin{figure}[tb]
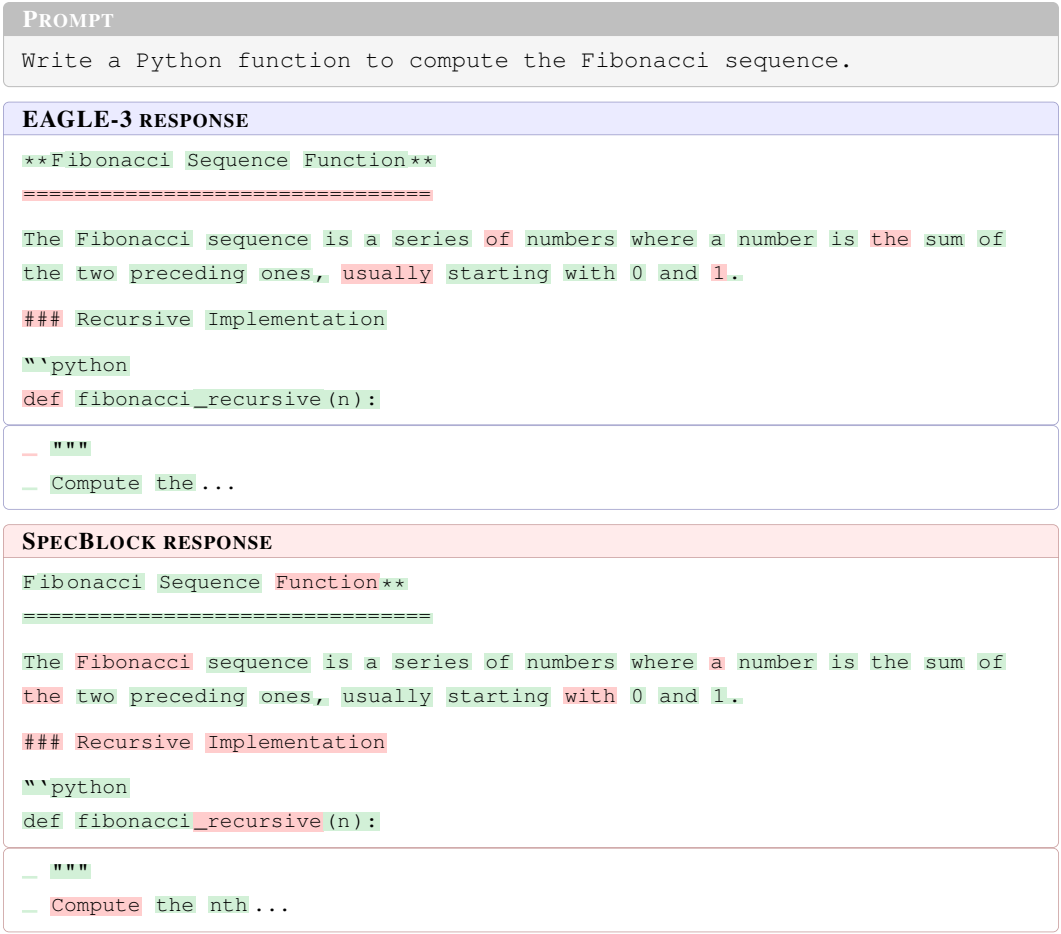

\centering
\begin{minipage}{\linewidth}
\begin{tcolorbox}[
  colback=gray!8, colframe=gray!50, boxrule=0.4pt,
  arc=2pt, left=4pt, right=4pt, top=3pt, bottom=3pt,
  title={\textsc{Prompt}}, fonttitle=\bfseries\small,
  breakable,
]
\ttfamily\small
Write a Python function to compute the Fibonacci sequence.
\end{tcolorbox}

\begin{tcolorbox}[colback=white, colframe=blue!30!gray!60, boxrule=0.3pt, title={\textsc{EAGLE-3 response}}, fonttitle=\bfseries\small, coltitle=black, colbacktitle=blue!8, arc=2pt, left=4pt, right=4pt, top=3pt, bottom=3pt, breakable]
\ttfamily\footnotesize\setlength{\fboxsep}{0.5pt}\setlength{\parindent}{0pt}\raggedright\sloppy\hyphenpenalty=10000\exhyphenpenalty=10000\linespread{1.35}\selectfont
\colorbox{acceptbg}{**}\colorbox{acceptbg}{F}\colorbox{acceptbg}{ib}\colorbox{acceptbg}{onacci} \colorbox{acceptbg}{Sequence} \colorbox{acceptbg}{Function}\colorbox{acceptbg}{**}\\
\colorbox{bonusbg}{================================}\\
\medskip
\colorbox{acceptbg}{The} \colorbox{acceptbg}{Fibonacci} \colorbox{acceptbg}{sequence} \colorbox{acceptbg}{is} \colorbox{acceptbg}{a} \colorbox{acceptbg}{series} \colorbox{bonusbg}{of} \colorbox{acceptbg}{numbers} \colorbox{acceptbg}{where} \colorbox{acceptbg}{a} \colorbox{acceptbg}{number} \colorbox{acceptbg}{is} \colorbox{bonusbg}{the} \colorbox{acceptbg}{sum} \colorbox{acceptbg}{of} \colorbox{acceptbg}{the} \colorbox{acceptbg}{two} \colorbox{acceptbg}{preceding} \colorbox{acceptbg}{ones}\colorbox{acceptbg}{,} \colorbox{bonusbg}{usually} \colorbox{acceptbg}{starting} \colorbox{acceptbg}{with} \colorbox{acceptbg}{0} \colorbox{acceptbg}{and} \colorbox{bonusbg}{1}\colorbox{acceptbg}{.}\\
\medskip
\colorbox{bonusbg}{\#\#\#} \colorbox{acceptbg}{Recursive} \colorbox{acceptbg}{Implementation}\\
\medskip
\colorbox{acceptbg}{```}\colorbox{acceptbg}{python}\\
\colorbox{bonusbg}{def} \colorbox{acceptbg}{fibonacci}\colorbox{acceptbg}{\_recursive}\colorbox{acceptbg}{(n}\colorbox{acceptbg}{):}\\
 \colorbox{bonusbg}{  } \colorbox{acceptbg}{"""}\\
 \colorbox{acceptbg}{  } \colorbox{acceptbg}{Compute} \colorbox{acceptbg}{the}\,\ldots
\end{tcolorbox}

\begin{tcolorbox}[colback=white, colframe=red!30!gray!60, boxrule=0.3pt, title={\textsc{SpecBlock response}}, fonttitle=\bfseries\small, coltitle=black, colbacktitle=red!8, arc=2pt, left=4pt, right=4pt, top=3pt, bottom=3pt, breakable]
\ttfamily\footnotesize\setlength{\fboxsep}{0.5pt}\setlength{\parindent}{0pt}\raggedright\sloppy\hyphenpenalty=10000\exhyphenpenalty=10000\linespread{1.35}\selectfont
\colorbox{acceptbg}{F}\colorbox{acceptbg}{ib}\colorbox{acceptbg}{onacci} \colorbox{acceptbg}{Sequence} \colorbox{bonusbg}{Function}\colorbox{acceptbg}{**}\\
\colorbox{acceptbg}{================================}\\
\medskip
\colorbox{acceptbg}{The} \colorbox{bonusbg}{Fibonacci} \colorbox{acceptbg}{sequence} \colorbox{acceptbg}{is} \colorbox{acceptbg}{a} \colorbox{acceptbg}{series} \colorbox{acceptbg}{of} \colorbox{acceptbg}{numbers} \colorbox{acceptbg}{where} \colorbox{bonusbg}{a} \colorbox{acceptbg}{number} \colorbox{acceptbg}{is} \colorbox{acceptbg}{the} \colorbox{acceptbg}{sum} \colorbox{acceptbg}{of} \colorbox{bonusbg}{the} \colorbox{acceptbg}{two} \colorbox{acceptbg}{preceding} \colorbox{acceptbg}{ones}\colorbox{acceptbg}{,} \colorbox{acceptbg}{usually} \colorbox{acceptbg}{starting} \colorbox{bonusbg}{with} \colorbox{acceptbg}{0} \colorbox{acceptbg}{and} \colorbox{acceptbg}{1}\colorbox{acceptbg}{.}\\
\medskip
\colorbox{bonusbg}{\#\#\#} \colorbox{bonusbg}{Recursive} \colorbox{bonusbg}{Implementation}\\
\medskip
\colorbox{acceptbg}{```}\colorbox{acceptbg}{python}\\
\colorbox{acceptbg}{def} \colorbox{acceptbg}{fibonacci}\colorbox{bonusbg}{\_recursive}\colorbox{acceptbg}{(n}\colorbox{acceptbg}{):}\\
 \colorbox{acceptbg}{  } \colorbox{acceptbg}{"""}\\
 \colorbox{acceptbg}{  } \colorbox{bonusbg}{Compute} \colorbox{acceptbg}{the} \colorbox{acceptbg}{nth}\,\ldots
\end{tcolorbox}
\end{minipage}
\caption{Case study response with per-token acceptance shading.
{\setlength{\fboxsep}{1pt}\colorbox{acceptbg}{green token}} indicates a
token drafted by the drafter and accepted by the verifier;
{\setlength{\fboxsep}{1pt}\colorbox{bonusbg}{red token}} indicates a
bonus token sampled from the target after acceptance. Each response is
shown for the first 30 committed tokens, with a trailing $\ldots$
marking the omitted tail.}
\label{fig:case_response}
\end{figure}

Figure~\ref{fig:case_tree} visualizes the iter-$0$ draft tree of each drafter on the same prompt; node coloring shows which drafter forward produced each token. EAGLE-3 grows depth-by-depth, so reaching depth~$7$ costs seven sequential drafter forwards, with each forward shown in a different color, fwd~$1$ through fwd~$7$. SpecBlock pays only two forwards on the same prompt: block-$1$ emits all $K{=}4$ chain positions in one forward drawn horizontally, and block-$2$ then batches additional chains from rank-head--selected starts; the two blocks are colored differently. Both drafters land on the same accepted prefix \texttt{F\,ib\,onacci\,Sequence}. EAGLE-3 keeps drafting and the verifier walks three more depths, committing eight tokens including a bonus over seven forwards. SpecBlock stops at the block-$1$ chain and commits five tokens including a bonus over two forwards. The case makes the SpecBlock trade-off visible: SpecBlock trades a slightly shorter accepted run per iteration for an iteration that finishes in two drafter calls instead of seven.


\definecolor{tgreen}{RGB}{ 30, 130,  60}
\definecolor{tred}{RGB}{200,  40,  40}
\definecolor{tdim}{RGB}{140, 140, 140}
\definecolor{tbox}{RGB}{120, 120, 120}

\definecolor{pal1}{RGB}{200, 220, 245}
\definecolor{pal2}{RGB}{180, 215, 230}
\definecolor{pal3}{RGB}{200, 220, 200}
\definecolor{pal4}{RGB}{240, 230, 180}
\definecolor{pal5}{RGB}{250, 215, 165}
\definecolor{pal6}{RGB}{245, 190, 160}
\definecolor{pal7}{RGB}{235, 165, 160}

\newcommand{\acc}{\textcolor{tgreen}{$\checkmark$}}
\newcommand{\rej}{\textcolor{tred}{$\times$}}
\newcommand{\accbox}[1]{\textbf{\textcolor{tgreen}{#1}}}
\newcommand{\dimcomment}[1]{{\color{tdim}\itshape #1}}
\newcommand{\treerow}[1]{#1\\}
\newcommand{\fwbox}[2]{{\setlength{\fboxsep}{0.5pt}\colorbox{#1}{\strut #2}}}
\newcommand{\fwA}[1]{\fwbox{pal1}{#1}}
\newcommand{\fwB}[1]{\fwbox{pal2}{#1}}
\newcommand{\fwC}[1]{\fwbox{pal3}{#1}}
\newcommand{\fwD}[1]{\fwbox{pal4}{#1}}
\newcommand{\fwE}[1]{\fwbox{pal5}{#1}}
\newcommand{\fwF}[1]{\fwbox{pal6}{#1}}
\newcommand{\fwG}[1]{\fwbox{pal7}{#1}}
\newcommand{\accAfwA}[1]{\fwA{\textbf{\textcolor{tgreen}{#1}}}}
\newcommand{\accAfwB}[1]{\fwB{\textbf{\textcolor{tgreen}{#1}}}}
\newcommand{\accAfwC}[1]{\fwC{\textbf{\textcolor{tgreen}{#1}}}}
\newcommand{\accAfwD}[1]{\fwD{\textbf{\textcolor{tgreen}{#1}}}}
\newcommand{\accAfwE}[1]{\fwE{\textbf{\textcolor{tgreen}{#1}}}}
\newcommand{\accAfwF}[1]{\fwF{\textbf{\textcolor{tgreen}{#1}}}}
\newcommand{\accAfwG}[1]{\fwG{\textbf{\textcolor{tgreen}{#1}}}}

\begin{figure}[!tb]
\centering

\begin{tcolorbox}[
  enhanced, colback=white, colframe=tbox, boxrule=0.4pt,
  arc=2pt, left=10pt, right=10pt, top=4pt, bottom=4pt,
  width=\linewidth,
  title={\textbf{(a) EAGLE-3}},
  fonttitle=\bfseries\small, coltitle=black, colbacktitle=blue!8,
]
{\ttfamily\footnotesize
\begin{alltt}
Write a Python function to compute the Fibonacci sequence. <|eot|> \textbackslash n\textbackslash n

root
└── \accAfwA{[F] \acc}
    ├── \accAfwB{[ib] \acc}
    │   ├── \accAfwC{[onacci] \acc}
    │   │   ├── \accAfwD{[ Sequence] \acc}
    │   │   │   ├── \accAfwE{[ Function] \acc}
    │   │   │   │   ├── \accAfwF{[**\textbackslash n] \acc}
    │   │   │   │   │   ├── \accAfwG{[================] \acc}
    │   │   │   │   │   └── \fwG{[The]}
    │   │   │   │   ├── \fwF{[**\textbackslash n\textbackslash n]}
    │   │   │   │   ├── \fwF{[ to]}
    │   │   │   │   └── \fwF{[:]}
    │   │   │   ├── \fwE{[ Calculator]}
    │   │   │   ├── \fwE{[ Generator]}
    │   │   │   ├── \fwE{[ with]}
    │   │   │   └── \dimcomment{... (9 more siblings)}
    │   │   ├── \fwD{[Sequence]}
    │   │   ├── \fwD{[ Series]}
    │   │   └── \fwD{[ Function]}
    │   └── \fwC{[)]}
    └── \fwB{[(n]}
        └── \fwC{[)]}
\end{alltt}}
\end{tcolorbox}

\vspace{4pt}

\begin{tcolorbox}[
  enhanced, colback=white, colframe=tbox, boxrule=0.4pt,
  arc=2pt, left=10pt, right=10pt, top=4pt, bottom=4pt,
  width=\linewidth,
  title={\textbf{(b) SpecBlock}},
  fonttitle=\bfseries\small, coltitle=black, colbacktitle=red!8,
]
{\ttfamily\footnotesize
\begin{alltt}
Write a Python function to compute the Fibonacci sequence. <|eot|> \textbackslash n\textbackslash n

root
├── \accAfwA{[F] \acc} ── \accAfwA{[ib] \acc} ── \accAfwA{[onacci] \acc} ── \accAfwA{[ Sequence] \acc}
│    │        │         │              │
│    │        │         │              ├── \fwG{[ using]}
│    │        │         │              │   ├── \fwG{[ Python]}
│    │        │         │              │   └── \fwG{[The]}
│    │        │         │              ├── \fwG{[ in]}
│    │        │         │              ├── \fwG{[**\textbackslash n]}
│    │        │         │              └── \dimcomment{... (8 more siblings)}
│    │        │         ├── \fwA{[ Sequence]}
│    │        │         ├── \fwA{[ using]}
│    │        │         └── \dimcomment{... (4 more siblings)}
│    │        ├── \fwA{[ Fibonacci]} ── \fwA{[onacci]} ── \fwA{[ Sequence]}
│    │        ├── \fwA{[:**]}
│    │        ├── \fwA{[ Implementation]} ── \fwA{[ of]}
│    │        └── \dimcomment{... (9 more siblings)}
│    └── \dimcomment{...}
├── \fwA{[ Fibonacci]} ── \fwG{[onacci]} ── \fwG{[ Sequence]}
├── \fwA{[The]} ── \fwG{[ Fibonacci]} ── \fwG{[ Sequence]}
├── \fwA{[What]} ── \fwG{[ is]} ── \fwG{[ the]} ── \fwG{[ Fibonacci]} ── \fwG{[ Sequence]}
├── \fwA{[Python]}
└── \fwA{[Overview]}
\end{alltt}}
\end{tcolorbox}

\caption{Per-iteration draft tree for the prompt
\texttt{Write a Python function to compute the Fibonacci sequence.}
EAGLE-3 grows depth-by-depth at one drafter forward per depth; each of
the seven forwards is shown in a different color, fwd 1 through fwd 7.
SpecBlock reaches a comparable accepted prefix in only two forwards,
with block-1 and block-2 shown in two different colors. Tokens marked
\acc{} are on the verifier-walked accept path; ``$N$ more siblings''
counts candidates omitted for clarity.}
\label{fig:case_tree}
\end{figure}

\section{Limitations}
\label{app:limitations}

SpecBlock builds its verifier tree using the rank head's per-position prediction, which decides how many sibling alternatives each position carries and whether the position starts a later block. The shape of the tree therefore depends on how accurate this prediction is. The four-bucket classifier already does better than a uniform tree of the same node budget, as the rank-head ablation in Table~\ref{tab:ablations} confirms, but it is not perfectly accurate. On a non-trivial fraction of positions the prediction misses by one or two buckets, so the position ends up with too few siblings when the target token sits far down the drafter's distribution, and too many when the drafter is already confident. A finer-grained or more accurate rank head would let SpecBlock spend its verifier budget more tightly.

The block width $K{=}4$ is decided at training time and cannot be changed at inference, because the layer-wise shift mechanism is built around a specific $K$. The number of iterative blocks $M$, in contrast, can be extended naturally at inference. A drafter trained with $M{=}3$ continues to work at $M{=}4$, since each additional block simply reuses the same drafter forward on a new starting point. In our experiments we use $M{=}2$ across deployments rather than searching $M$ per workload. Both $K$ at training and $M$ at inference interact with the cost ratio between the target and the drafter. A much larger target makes each verifier call expensive and rewards a larger $K$ or deeper block stacking. When the two costs are closer, smaller values save drafter time without losing much. Workloads whose acceptance distribution differs noticeably from our training mix may also prefer values we did not select.

\end{document}